\documentclass[transmag]{IEEEtran}

\usepackage{amsmath,epsfig}
\usepackage{amssymb}
\usepackage{bm}
\usepackage{cite}
\usepackage{url}
\usepackage{booktabs}
\usepackage{multirow}
\usepackage{makecell}
\usepackage{xcolor}
\usepackage{colortbl}
\usepackage{fancyhdr}

\usepackage{calc}
\newlength{\maxmin}

\usepackage{xspace} 
\usepackage{algorithm,algpseudocode}
\usepackage{comment}
\usepackage{mathtools}
\usepackage{empheq}
\usepackage{color}
\definecolor{mypeach}{RGB}{249,231,223}
\definecolor{mygreen}{RGB}{0,109,117}
\usepackage[most]{tcolorbox}
\tcbuselibrary{theorems}

\newcommand{\caseif}{\text{if }} 

\DeclareRobustCommand{\resnetfifty}{ResNet\nobreakdash-50\xspace}

\DeclareMathOperator*{\argmin}{arg\,min}

\begin{document}

\title{Lightweight Compression of Intermediate Neural Network Features for Collaborative Intelligence}

\author{
Robert A. Cohen, \IEEEmembership{Senior Member,~IEEE},
Hyomin Choi, \IEEEmembership{Student Member,~IEEE},\\
and Ivan V. Baji\'{c} \IEEEmembership{Senior Member,~IEEE}\\
\thanks{Robert A. Cohen, Hyomin Choi, and Ivan V. Baji\'{c} are with
the School of Engineering Science,
Simon Fraser University, BC, V5A 1S6, Canada.
(email: [robert\_cohen, chyomin, ibajic]@sfu.ca).}
}

\IEEEtitleabstractindextext{
\begin{abstract}
In collaborative intelligence applications, part of a deep neural network (DNN) is deployed on a lightweight device such as a mobile phone or edge device, and the remaining portion of the DNN is processed where more computing resources are available, such as in the cloud. This paper presents a novel lightweight compression technique designed specifically to quantize and compress the features output by the intermediate layer of a split DNN, without requiring any retraining of the network weights. Mathematical models for estimating the clipping and quantization error of ReLU and leaky-ReLU activations at this intermediate layer are developed and used to compute optimal clipping ranges for coarse quantization. We also present a modified entropy-constrained design algorithm for quantizing clipped activations. When applied to popular object-detection and classification DNNs, we were able to compress the 32-bit floating point intermediate activations down to 0.6 to 0.8 bits, while keeping the loss in accuracy to less than 1\%. When compared to HEVC, we found that the lightweight codec consistently provided better inference accuracy, by up to 1.3\%. The performance and simplicity of this lightweight compression technique makes it an attractive option for coding an intermediate layer of a split neural network for edge/cloud applications.
\end{abstract}
}

\maketitle

\enlargethispage{-30pt}

\fancypagestyle{firstpage}
{
    \fancyhf{}
    \fancyhead[C]{
    {\scriptsize{Copyright \copyright 2020 IEEE. Personal use of this material is permitted. Permission from IEEE must be obtained for all other uses, in any current or future media, including reprinting/republishing this material for advertising or promotional purposes, creating new collective works, for resale or redistribution to servers or lists, or reuse of any copyrighted component of this work in other works.}
    \\}
    \vspace{12pt}
    \textcolor{mygreen}{For the published open-access version of this paper, please visit: https://doi.org/10.1109/OJCAS.2021.3072884}}
    \setlength{\headheight}{49pt}
    \renewcommand{\headrulewidth}{0.0pt}
}

\thispagestyle{firstpage}


\section{Introduction}
\label{sec:intro}
With the increasing ubiquity of intelligent applications in our daily lives, machine learning and artificial neural networks are rapidly finding their way into a wide range of systems and devices, from large-scale cloud computing systems all the way down to handheld and even miniature implanted devices. As the amount of data captured and generated from devices such as smartphones and Internet-of-Things (IoT) sensors, the deep neural networks (DNNs) needed to accomplish the desired tasks can be too large or complex to realize entirely within a device. For devices located at the edge of a network,
lightweight and mobile-friendly architectures~\cite{Chen2019,Lane2018,Tan_2019_CVPR} can facilitate the implementation of these DNNs. When a DNN that performs real-time inference or other compute-intensive operations is too complex to realize fully on an edge device, a collaborative intelligence~\cite{Kang2017,Bajic_etal_ICASSP21} paradigm can be used to split the DNN so that the bulk of the computations can be performed in the cloud. In this case, a subset of layers of the DNN is computed inside the edge device, and then the output of the last layer on the device is signaled to the cloud, to be used as input to the remaining layers of the DNN. The ideal location to split the DNN can be determined by examining both the available computational resources in the edge device and the size of the data such as feature tensors that need to be signaled~\cite{Kang2017}.

The amount of data or the size of feature tensors generated by hidden layers in a DNN can be quite inflated compared to the size of the DNN's input or output data.
If a network is split at a hidden layer, then data reduction and compression methods are needed to make the signaling of feature tensors practical.
Typical approaches for data reduction in neural networks include network optimization methods such as factorization~\cite{zhang2015efficient}, pruning~\cite{zhuang2018discrimination}, and network parameter quantization~\cite{Mishra2017_WRPN,jacob2018quantization}. 
The goal of many of the neural network quantization methods is to reduce the size of the data, complexity of operations, and amount of memory so that the DNN can operate more efficiently on a device.
In the context of collaborative intelligence, an additional goal is to process and efficiently compress the features produced by the DNN's front-end, so that they can be signaled between an edge device and the cloud.

\enlargethispage{-30pt}

In our earlier work~\cite{Cohen_quantcode_ICME2020}, we presented a lightweight compression method that is well-suited for coding the output of a split DNN in edge-based devices. This method uses simple and very coarse scalar quantization along with clipping, binarization, and entropy coding to compress the activations, without needing any retraining of network weights, i.e., post-training quantization. All the clipping ranges in that paper were determined empirically. In this paper, we introduce mathematical models for the distributions of feature tensors output by a leaky rectified linear unit (ReLU)~\cite{Maas2013RectifierNI}. We use these models to obtain closed-form expressions for clipping and quantization error, from which we can obtain optimal clipping values. We show how well these models estimate the error, and then we compare the overall neural network performance using these models to the performance obtained when using empirically determined clipping ranges. We also provide additional detailed experimental results for the networks used in this paper.

In Section~\ref{sec:related_work} we describe prior related works on quantization and compression in neural networks. In Section~\ref{sec:lightweight}, we present the lightweight codec, models for optimal clipping and quantization, discussions and illustrations of how the models behave, methods for improving the system's compression efficiency, and computational complexity comparisons. Section~\ref{sec:experiments} presents experimental results and comparisons between empirical and model-based performance, followed by conclusions in Section~\ref{sec:conclusions}.


\section{Related work}
\label{sec:related_work}
Methods for quantizing or compressing neural networks weights or feature tensors generally fall into two categories: quantization-aware training and post-training quantization. In quantization-aware training, the network weights are trained with quantization applied or simulated in floating point, often in the forward pass, with higher-precision arithmetic used during stochastic gradient descent to assist with convergence of the training process. For example, Jacob, et al.~\cite{jacob2018quantization} uses that approach to quantize weights and activations to 8 bits, while quantizing bias parameters to 32 bits.
Mishra, et al.~\cite{Mishra2017_WRPN} find that when increasing the batch size during training of a low-precision network, the proportion of memory used by activation maps increases significantly. To reduce the precision of
weights and activations while maintaining the overall accuracy of the network, they insert additional filter maps into each layer. They find that for networks such as ResNet\nobreakdash-34~\cite{He2015DeepRL} and others that near full-precision performance can be achieved with 4-bit activations and 2-bit weights. In~\cite{Banner2018_8BitTraining}, weights, activations, and some of the gradient and back-propagation computations are quantized to 8 bits, along with a scaling modification made to batch normalization.
In~\cite{Choi2019_2bit}, networks are trained with weights and activations quantized to two bits.
Good performance is achievable even with extremely coarse quantization, as shown in~\cite{Hubara2016_BNN,Rastegari2016_XNOR} where filters and the inputs to convolutional layers are one-bit values.

For quantizing neural networks that have already been trained, including those trained without considering quantization, post-training quantization can be applied during inference. If this quantization is sufficiently fine or if the quantizer design method is tailored to the characteristics of the weights or activations being quantized, then the performance of this inference can become quite good, and in some cases can match the performance of a system designed using quantization-aware training or even no quantization at all.
For example,~\cite{Krishnamoorthi2018_QuantizingDC} applies straightforward 8-bit quantization to weights and activations while maintaining nearly the same accuracy as obtained with floating point, and with 4-bit quantization, a 5\% loss in accuracy is observed. Two approaches that deal with how to quantize floating-point values that have a high dynamic range are outlier-aware quantization and clipping. In~\cite{Park2018a,Park2018b}, quantization of weights and activations to 4 bits is achieved by using a coarse quantizer for 97\% of the values and a finer quantizer for the remaining outliers. OCS~\cite{Zhao2019_Quantization} instead splits a channel into two channels in which the weights and outputs are halved, which reduces the dynamic range of the outliers.
DFQ~\cite{Nagel2019} quantizes weights and activations to 8 bits by assuming that the inputs to the activations have a Gaussian distribution, so that a model can be used to equalize the dynamic range of the data being quantized, along with a correction to the bias introduced by quantization.
Clipping-based approaches are used in~\cite{banner2018_ACIQ,Banner2019_4bit,Cohen_quantcode_ICME2020}, in which activations or weights are clipped prior to quantization. Banner, et al.~\cite{banner2018_ACIQ,Banner2019_4bit} uses a piecewise linear model of symmetric or one-sided Laplace and Gaussian distributions to model weights and activations to determine optimal clipping values, along with a simple bias correction for weights. That method can be used to quantize neural networks layers to an average of four bits. For networks that use batch normalization, ZeroQ~\cite{Cai2019} obtains model parameters from synthetic data instead of from training or validation sets. The synthetic data is generated based upon minimizing a distortion metric that computes the respective differences between the means and variances of batch normalization and those of the reconstructed synthetic input. Pareto optimization is used to select the bit precision for each layer.
In~\cite{Cohen_quantcode_ICME2020}, empirical-based clipping is used to apply extremely coarse post-training quantization to activations, down to one bit. Fractional bit precisions are supported as well.

For collaborative intelligence applications, in which a DNN is split, additional compression is applied after quantization, so that the compressed activations can be efficiently signaled between a device and the cloud. In~\cite{dfc_for_collab_object_detection,eshratifar2019jointdnn}, feature tensors are quantized to 8 bits and then converted to tiled images to be compressed by off-the-shelf codecs. However, conventional image and video codecs are tailored to coding camera-captured images, whereas images formed using feature tensors exhibit significantly different characteristics. To address this discrepancy, the codec of~\cite{Choi2018NearLosslessDF} is tailored according to the feature tensor statistics so that lossless coding can be applied to 8-bit pre-quantized features. Lossy compression can also be used to further compress the activations output at a split layer.
In~\cite{eshratifar2019towards, eshratifar2019bottlenet}, a small auto-encoder-like neural network is introduced between the edge front-end and the cloud back-end, to reduce the dimensions of the signaled feature tensor. However, this approach requires end-to-end (re)training of the original DNN to minimize accuracy loss caused by the dimension reduction. In~\cite{Choi_BaF_2020}, the dimension of the feature tensor is reduced using a small neural network in a process called Back-and-Forth (BaF) prediction, which does not require retraining of the original DNN. The reduced-dimension tensor is then further quantized to 6 bits and compressed with a state-of-the-art lossless codec.

To achieve high compression efficiency without requiring the complexity of an off-the-shelf image codec, our previous work~\cite{Cohen_quantcode_ICME2020} applies clipping and very coarse post-training quantization, e.g., up to two bits, to the output activations, followed by binarization and entropy coding to generate a compressed bit-stream. The optimal clipping values in~\cite{Cohen_quantcode_ICME2020} were determined empirically. Due to the asymmetry of the distributions of feature tensor values at our split layers and the extreme coarseness of our quantizers, the assumptions of symmetry and uniformity used for the piecewise model of~\cite{banner2018_ACIQ,Banner2019_4bit} do not apply. The purpose of this paper is to extend the work of~\cite{Cohen_quantcode_ICME2020} by developing a mathematical model of the feature tensors output by a leaky ReLU activation function whose input is asymmetric; using this model to estimate clipping and quantization error of the activations; determining how these error estimates behave with extremely coarse quantization; and applying these models to determine optimal clipping ranges for quantization. We also review the complete lightweight compression system, which includes an entropy-constrained quantizer design algorithm modified to pin the outermost quantizer reconstruction levels so that the dynamic range of clipped activations is preserved when decoded.


\section{Modeling and lightweight compression of feature tensors}
\label{sec:lightweight}
\begin{figure}[tb]
    \centering
    \includegraphics[width=0.48\textwidth,viewport=2.391047 126.719996 553.571983 538.217984,clip]{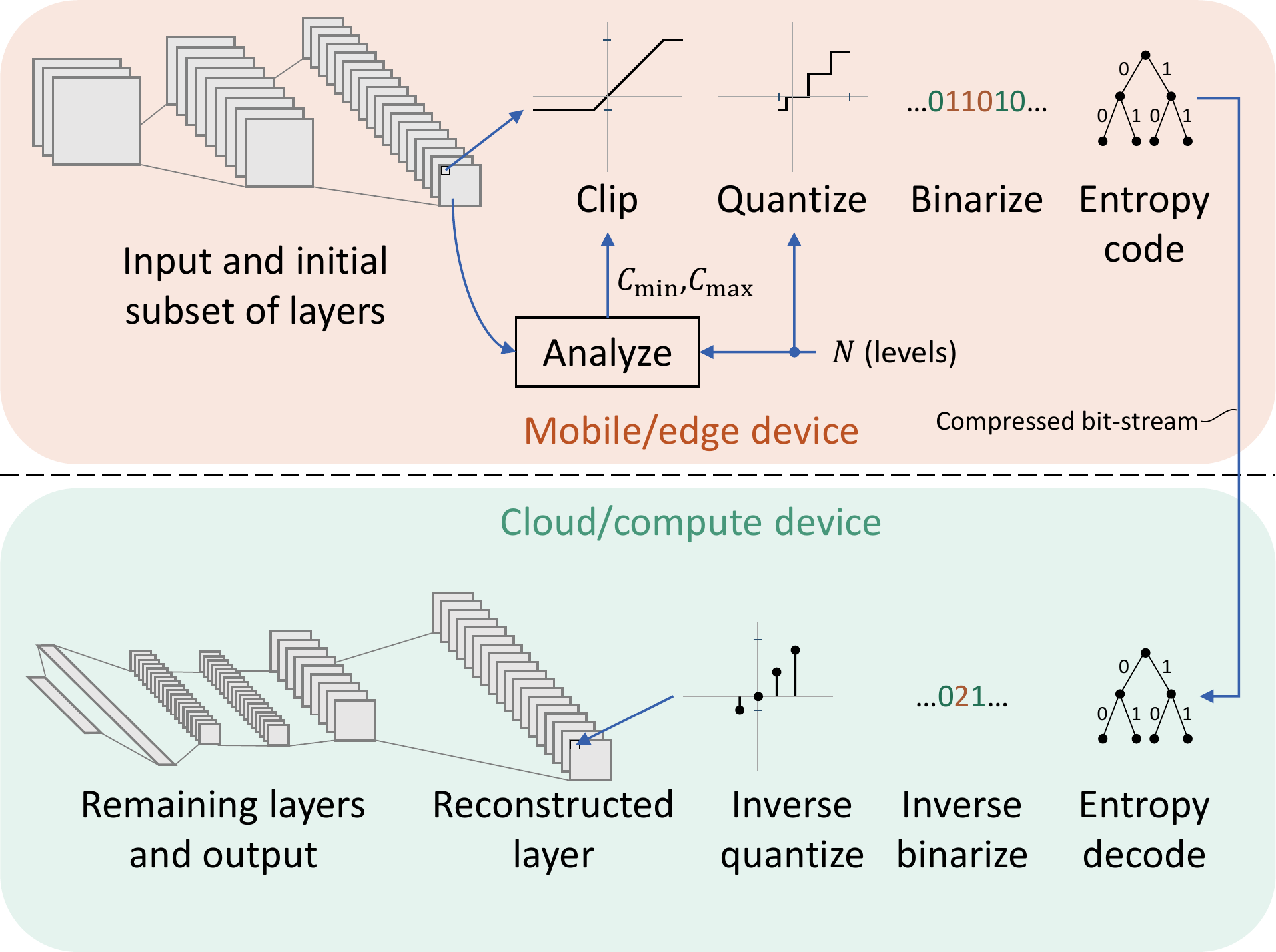}
    \caption{Lightweight compression system overview.}
    \label{fig:system}
\end{figure}

Fig.~\ref{fig:system} shows an overview of the proposed lightweight compression method used in a collaborative intelligence environment.
The first several layers of a DNN are processed on a mobile or edge device. A typical layer includes convolutions, batch normalization, and an activation function. The outputs of this subset of layers are signaled to the cloud or to another computing device for processing by the remaining DNN layers. To efficiently transmit the activations or feature tensors from the first portion of the DNN, data compression is needed. Because a mobile or edge device may have limitations on computing complexity or available energy, a low-complexity compression method is preferred, if it does not significantly impact on overall accuracy of the DNN. The lightweight compression system illustrated here uses simple operations including clipping and very coarse memoryless scalar quantization to represent the feature tensors using a small set of quantized symbols. These symbols are binarized and entropy coded to further compress the data, and the compressed bit-stream is transmitted to a different computing platform to be decoded and used as input to the remaining layers of the DNN.
The minimum and maximum clipping values $c_\mathrm{min}$ and $c_\mathrm{max}$ can be determined empirically or via a model-based analysis based on the sample mean and variance of data to be compressed.

It is well known that 8-bit post-training quantization of 32-bit floating point neural network activations can be applied with little effect on the overall accuracy of the network~\cite{Vanhoucke2011}. Likewise, the same quantization can be applied to the intermediate layer where a DNN is split for a collaborative intelligence application. When quantizing without special considerations to four or fewer bits, however, performance can be significantly degraded. For example, if we split \resnetfifty~\cite{He2015DeepRL} at layer 21 and apply 3-bit uniform quantization to the activations output by that layer, then the Top\nobreakdash-1 classification accuracy over the ImageNet ILSVRC2012~\cite{imagenet2015} validation data set goes from 75.8\% without this quantization to 59.7\% with quantization.

Most if not all this performance loss can be mitigated simply by clipping the activations before quantization, without any retraining of the DNN weights. In this section, we first examine the effects of clipping on the overall network performance. We then present an analytic model for obtaining optimal clipping parameters prior to coarse quantization, and we examine in detail how well this model works for different levels of coarse quantization. We also review the methods used to achieve further compression, such as binarization and a modified entropy-constrained design process for quantizing clipped values. 

\subsection{Effects of clipping}
\label{sec:clipping}
\begin{figure}[tb]
    \centering
    \begin{minipage}[b]{\columnwidth}
        \centering
        \includegraphics[width=.9\textwidth,viewport=7.740000 7.794000 701.243979 413.081987,clip]{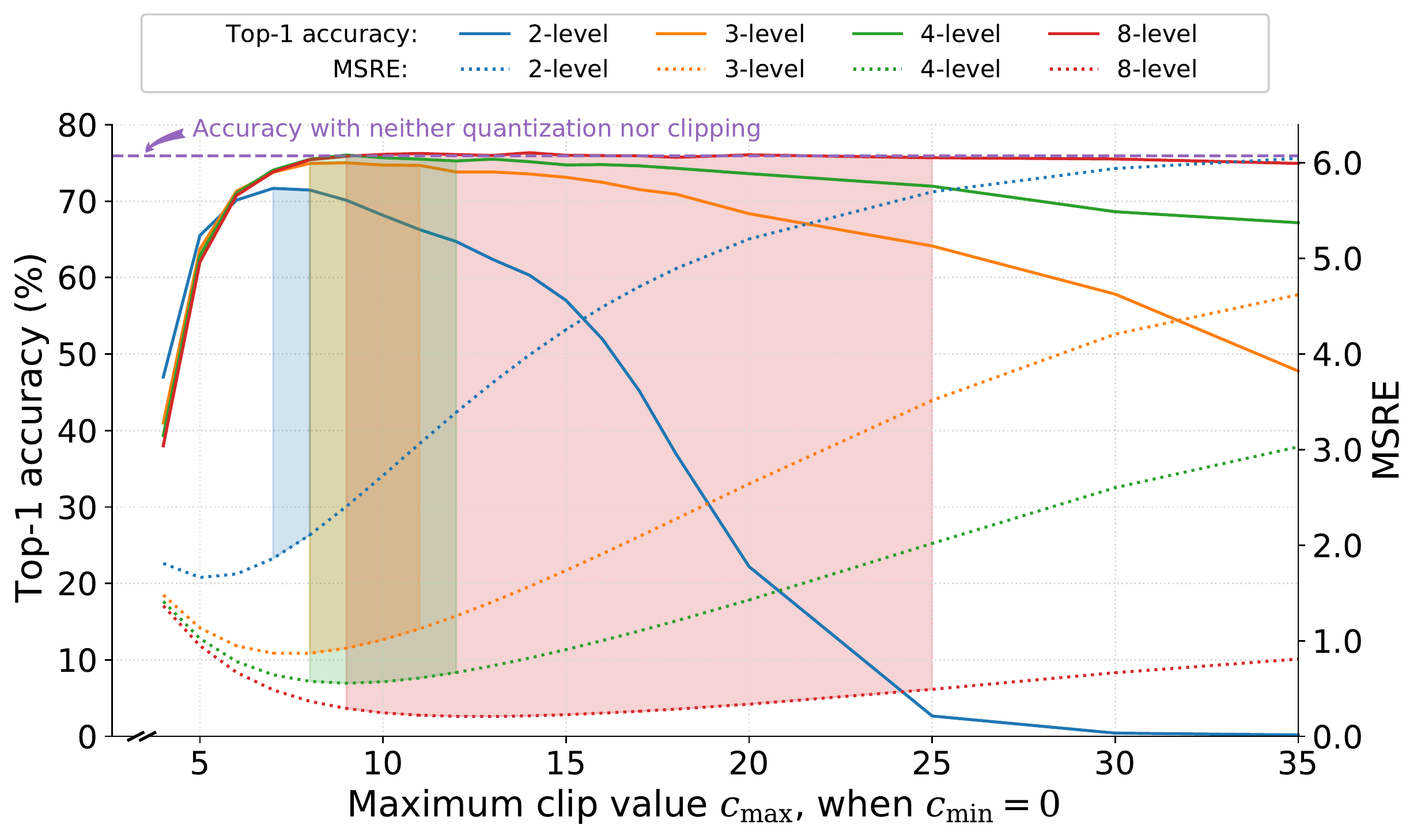}\\
        (a) \resnetfifty
    \end{minipage}
    \vspace{6pt}
    \begin{minipage}[b]{\columnwidth}
        \centering
        \includegraphics[width=.9\textwidth,viewport=7.740000 7.794000 701.243979 413.081987,clip]{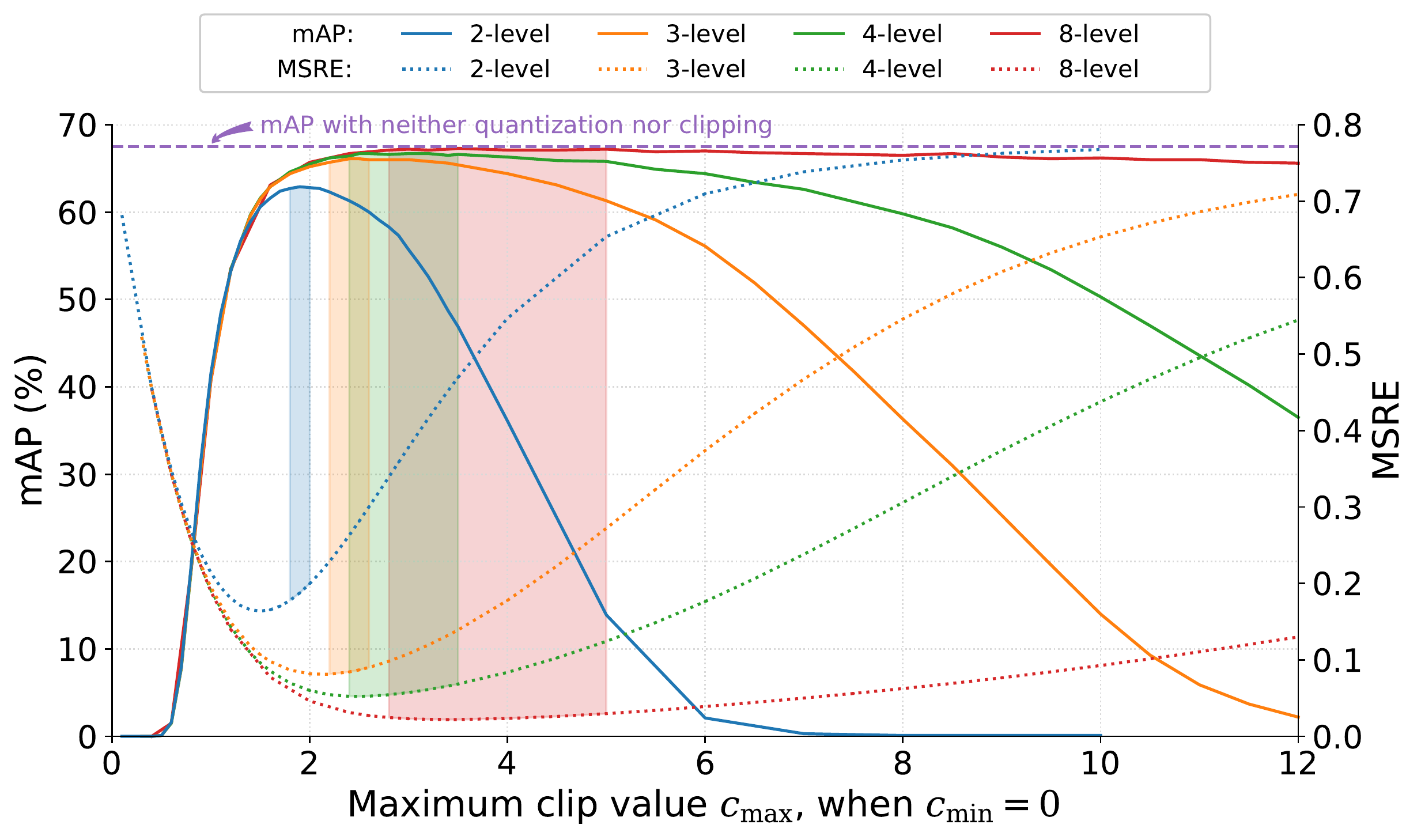}\\
        (b) YOLOv3
    \end{minipage}
    \vspace{6pt}
    \begin{minipage}[b]{\columnwidth}
        \centering
        \includegraphics[width=.9\textwidth,viewport=7.200000 7.794000 699.245979 413.081987,clip]{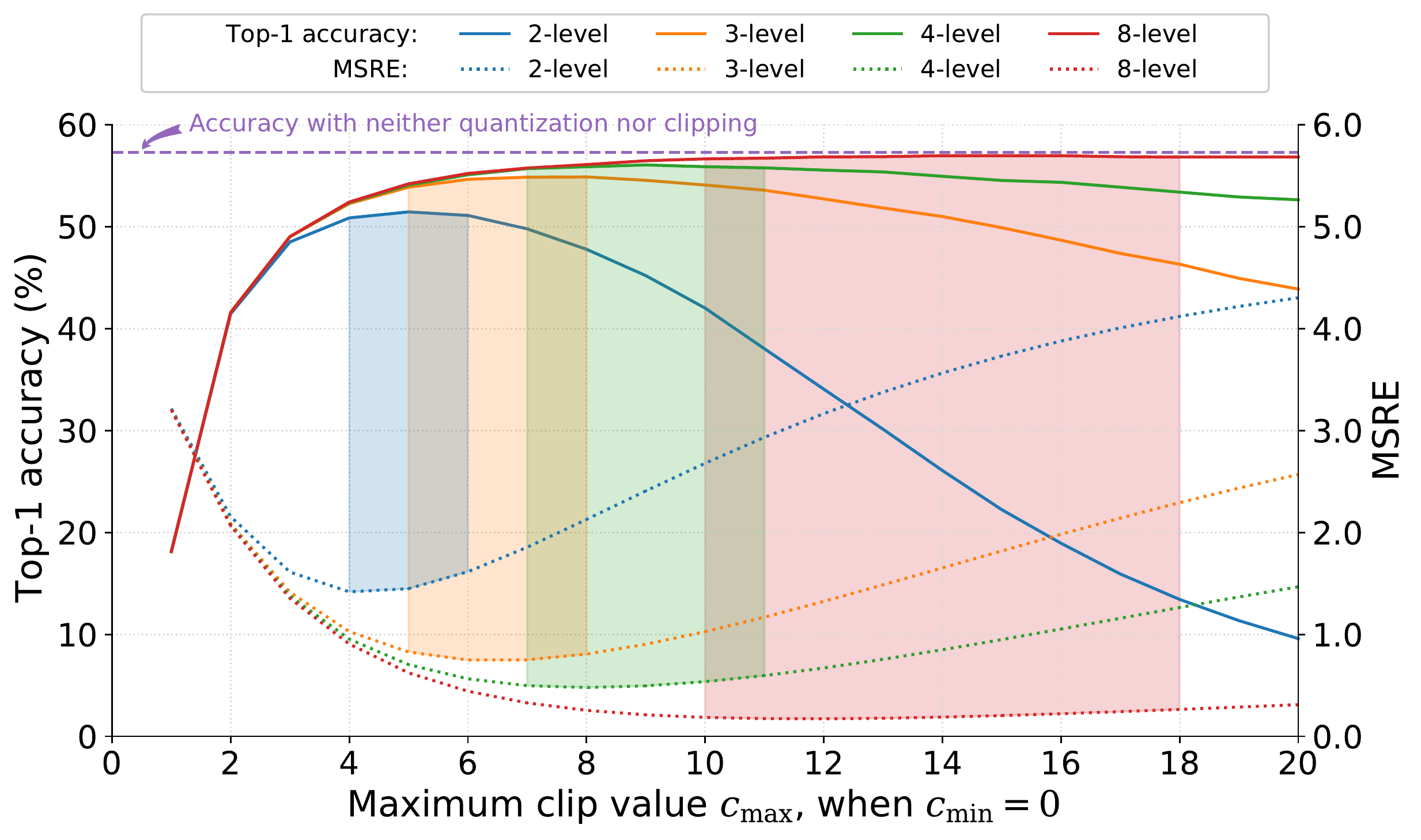}\\
        (c) AlexNet
    \end{minipage}\hfill    
    \caption{Effects of clipping (a) layer 21 activations in \resnetfifty, (b) layer 12 activations in YOLOv3, and (c) layer 4 activations in AlexNet. Shaded areas represent ranges of maximum clipping values capable of achieving peak network accuracy or precision.}
    \label{fig:clip_range_resnet_yolo_alexnet}
\end{figure}

Fig.~\ref{fig:clip_range_resnet_yolo_alexnet}(a) shows the effects that clipping and coarse quantization have on the Top-1 classification accuracy of the \resnetfifty classification network when averaged over 5k images from the ImageNet ILSVRC2012 validation data set. Here, the activations at the output of layer 21 are clipped (clamped) to be between $c_\mathrm{min} = 0.0$ and $c_\mathrm{max}$. Layer 21 of \resnetfifty corresponds to the output of the shortcut connection and element-wise addition applied to the output of the second (out of four) residual blocks in the \texttt{conv3\_x} layer shown in~\cite[Table~1]{He2015DeepRL}.
If we instead use AlexNet~\cite{alexnet} with the ImageNet validation set and apply clipping and quantization to the output of layer 4, we obtain Fig.~\ref{fig:clip_range_resnet_yolo_alexnet}(c). Layer 4 of AlexNet corresponds to the convolutional layer immediately after the second maxpool layer shown in~\cite[Figure 2]{alexnet}.
A similar plot is shown in Fig.~\ref{fig:clip_range_resnet_yolo_alexnet}(b) for the mean Average Precision (mAP) of the YOLOv3~\cite{Redmon2018_yolov3} object detection network, when run on the COCO 2017~\cite{COCO} validation data set with Intersection-over-Union (IoU) threshold set to 0.5. In this case, the output of layer 12 is clipped and quantized. Layer 12 of YOLOv3 corresponds to the output of the convolution just before the first $8\times$ group of residual blocks shown in~\cite[Table~1]{Redmon2018_yolov3}. We cut the networks at these layers because generally for collaborative intelligence applications, a subset of the neural network is implemented on a lightweight device. Therefore, the size of this subset is usually much smaller than the portion of the network implemented in the cloud. However, we do not want to cut the network too early, as the size of the feature tensor typically grows rapidly in the first few layers. Additionally, we do not want to cut across many data paths at once. For example, the feature tensors in YOLOv3 rapidly expand by a factor of over ten times, and it is not until layer 12 until the feature tensor size comes back down to the same order of magnitude as the input.

Each clipped activation value, denoted as $x_\mathrm{clp}$, is processed by an $N$-level quantizer as follows:
\begin{equation}
\label{eq:clipquant}
    Q(x_\mathrm{clp}) =  \mathrm{round}\left(\frac{x_\mathrm{clp} - {c}_\mathrm{min}}{{c}_\mathrm{max} - {c}_\mathrm{min}}
    \cdot (N-1)\right) \, ,
\end{equation}
where $\mathrm{round(\cdot)}$ rounds away from zero for halfway cases. Note that unlike related literature that focuses on reduced bit-depth architectures, our $N$ does not need to be a power of two, as the purpose of quantization is for compression and subsequent transmission or storage in a bit-stream. The mean-square reconstruction error (MSRE) $\mathrm{E}\left[(x-{\hat x}_{\mathrm{clp}})^2\right]$ computed between an unmodified activation $x$ and the inverse-quantized clipped activation is shown using dotted lines.
For 8-level (3-bit) quantization of \resnetfifty layer 21 activations, Fig.~\ref{fig:clip_range_resnet_yolo_alexnet}(a) shows that peak Top-1 accuracy is achieved over a range of ${c}_\mathrm{max}$ values between roughly 9.0 and 25.0. This range is indicated by the correspondingly colored shaded region. As the number of quantization levels is decreased, the optimal ${c}_\mathrm{max}$ decreases, as does the range of ${c}_\mathrm{max}$ values that achieves peak performance. With 1-bit (2-level) quantization, the optimal range is quite narrow. When the quantization is not extremely coarse, e.g. 8-level (3-bit) or higher, the minimum MSRE generally coincides with the peak accuracy of \resnetfifty and peak mAP performance of YOLOv3, as can be seen in the plot where the minimum of the MSRE curve falls within the shaded region corresponding to the maximum network accuracy. Earlier works, e.g.~\cite{banner2018_ACIQ,Banner2019_4bit} have leveraged this behavior to model the quantization error to select the optimal clipping range for all activations in a DNN. However, it is evident from these prior works and explicitly stated by the authors that deviations from the models occur with extremely coarse quantization, e.g. corresponding to 2-bit (4-level) and below in this example. We can see in Fig.~\ref{fig:clip_range_resnet_yolo_alexnet}(a) for \resnetfifty that the optimal ${c}_\mathrm{max}$ for 2-level (1-bit) quantization is approximately 7.0, whereas the minimum MSRE occurs near ${c}_\mathrm{max}=5.0$. Similar behavior is exhibited in Fig.~\ref{fig:clip_range_resnet_yolo_alexnet}(b) for YOLOv3.
Thus, choosing ${c}_\mathrm{max}$ based on minimizing MSRE can result in a potential loss in accuracy of several percent when $N$ is small. Nonetheless, it is still worthwhile examining an analytical model for estimating the optimal clipping ranges, given that the model may still be useful for some values of $N$, and because empirically determining optimal ranges increases the complexity of the design process.

\subsection{Model for computing optimal clipping ranges for activations}
\label{sec:model}
\begin{figure}[tb]
    \begin{minipage}[b]{.42\columnwidth}
        \centering
        \includegraphics[width=\columnwidth,viewport=62.241115 489.077985 306.489014 735.407978,clip]{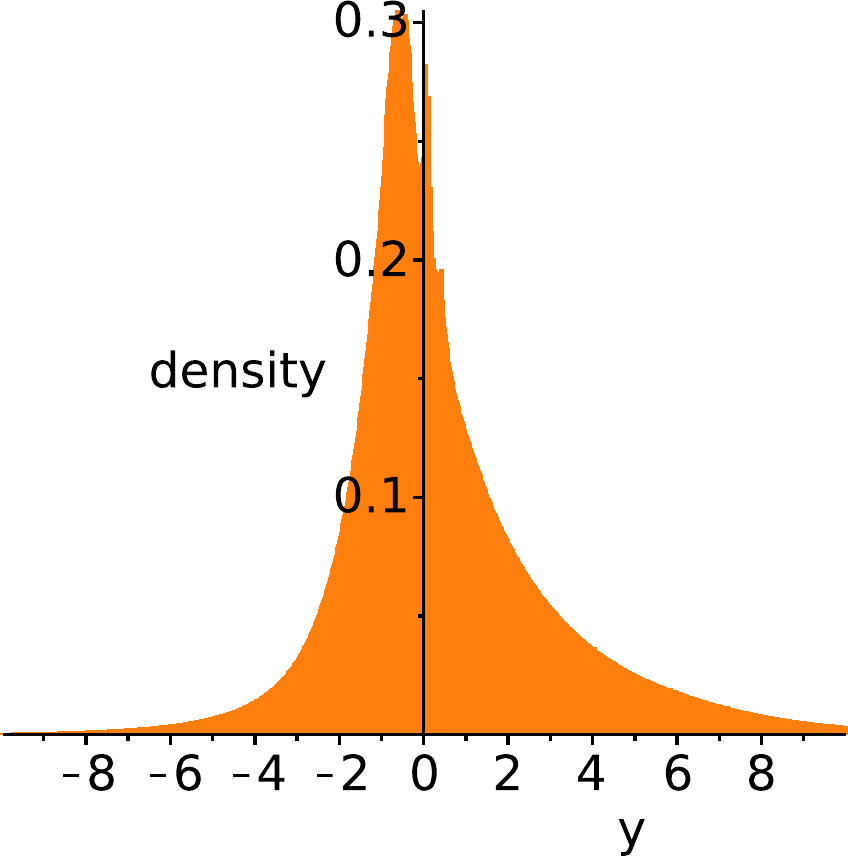}\\
        (a)
    \end{minipage}\hfill
    \begin{minipage}[b]{.42\columnwidth}
        \centering
        \includegraphics[width=\columnwidth,viewport=3.006000 1.800000 111.257997 110.573997,clip]{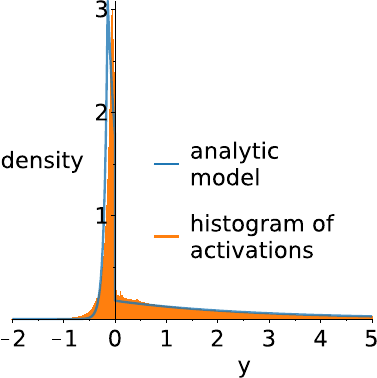}\\
        (b)
    \end{minipage}\hfill    
    \caption{Distribution of feature tensor values for layer 21 of \resnetfifty (a) before leaky ReLU, (b) after leaky ReLU including plot of the PDF from the analytic model.}
    \label{fig:hist_resnet}
\end{figure}

In a collaborative intelligence system where a neural network is split, we may only have access to the activations or feature tensors output by the front-end of the network. We would like to measure the statistical properties of that output to determine optimal clipping values. Prior works such as~\cite{banner2018_ACIQ,Banner2019_4bit} assume a Laplace or Gaussian model for the distribution of feature tensors, and for when a rectified linear unit (ReLU) is used for the activation function, a single-sided Laplace, i.e. exponential distribution is used as a model on the assumption that all negative values are rectified. For networks such as \resnetfifty, however, leaky ReLU is used for the activation function, in which negative values are preserved at a smaller scale. Fig.~\ref{fig:hist_resnet} shows the distribution of feature tensor elements immediately before and after leaky ReLU at the output of layer 21 of \resnetfifty when run on the full ImageNet ILSVRC2012 validation data set of 50k images. Due in part to the scaling of negative values in the earlier layers, we can see that the distribution of the data input to the layer 21 activation function is skewed and has a peak not located at zero. The output of layer 21 thus results from a leaky ReLU operation applied to an asymmetric distribution, rather than to a symmetric and zero-mean distribution, as assumed in earlier works.

For developing a model to compute optimal clipping ranges at the output of a leaky ReLU activation function, we model the input to the activation function as having an asymmetric Laplace distribution~\cite{Kozubowski2000} with probability density function (PDF)
\begin{equation}
f_\mathrm{L}(x) =
    \frac{\lambda}{\kappa+{\kappa}^{-1}} \cdot
    \begin{dcases}
        {e^{{\frac{\lambda\, \left( x-\mu \right) }{\kappa}}}} & \caseif x<\mu \\
        {e^{-\lambda\,\kappa\, \left( x-\mu \right) }} & \caseif x \geq \mu
    \end{dcases}
\end{equation}
where $\kappa$ is a constant that determines the asymmetry of the distribution, $\mu$ is the location of the peak of the distribution, and $\lambda > 0$. Note that, unlike the symmetric Laplace distribution, $\mu$ here is not the mean. For \resnetfifty, we use $\kappa = 0.5$ to obtain an asymmetric Laplace distribution that approximates the distribution of feature tensor elements input to leaky ReLU. The density function now becomes
\begin{equation}
f(x) =
    0.4 \lambda \cdot
    \begin{dcases}
        e^{2 \lambda(x-\mu)} & \caseif x<\mu \\
        e^{-0.5 \lambda(x-\mu)} & \caseif x \geq \mu \, .
    \end{dcases}
    \label{eq:relu_input_density}
\end{equation}
The \resnetfifty implementation that we use has a leaky ReLU activation function with a scaling factor of 0.1 for negative values, defined as follows:
\begin{equation}
    \mathrm{leaky\_ReLU}(x) =
    \begin{dcases}
    0.1 x & \caseif x < 0 \\
    \hphantom{-0.1} x & \caseif x \geq 0 \, .
    \end{dcases}
    \label{eq:leaky_relu}
\end{equation}
By applying~\eqref{eq:leaky_relu} to data having the distribution of~\eqref{eq:relu_input_density}, the output of leaky ReLU has the piecewise distribution
\setlength{\maxmin}{\widthof{$e^{-0.5 \lambda(10y-\mu)}$}-\widthof{$e^{-0.5 \lambda(y-\mu)}$}}
\begin{equation}
    f_\mathrm{Y}(y) = 
    \begin{dcases}
        \hphantom{0.}4 \lambda \cdot
        \begin{dcases}
            e^{2 \lambda(10y-\mu)} & \caseif y<0.1\mu \\
            e^{-0.5 \lambda(10y-\mu)} & \caseif y \geq 0.1\mu
        \end{dcases}   
        & \caseif y < 0
        \\
        0.4 \lambda \cdot
        \begin{dcases}
            e^{2 \lambda(y-\mu)} & \hspace{\maxmin} \caseif y<\mu \\
            e^{-0.5 \lambda(y-\mu)} & \hspace{\maxmin} \caseif y \geq \mu
        \end{dcases}
        & \caseif y \geq 0 \, .
    \end{dcases}
    \label{eq:relu_output_dist}
\end{equation}
We can see from Fig.~\ref{fig:hist_resnet}(a) that the peak of the histogram corresponds to a negative value, therefore $\mu <0$, and the mean of a random variable $Y$ having the PDF of~\eqref{eq:relu_output_dist} can be simplified to
\begin{equation}
\mathrm{E}\left[ Y \right] = 
0.1 \mu + \frac{1}{\lambda} \left[ \frac{3}{20} + {\left( \frac{6}{5} \right)}^2 e^{0.5 \lambda \mu} \right] \, ,
\label{eq:relu_output_mean}
\end{equation}
and the variance becomes
\begin{equation}
\begin{multlined}
\mathrm{E}\left[ {\left( Y - \mathrm{E}\left[ Y \right] \right)}^2 \right] = \\
    \frac{1}{\lambda^{2}}
    \left[
    \left(5.904 - 0.288 \lambda \mu \right) e^{0.5 \lambda \mu} - 2.0736 e^{\lambda \mu} + 0.0425
    \right] \, .
\end{multlined}
\label{eq:relu_output_var}
\end{equation}
By setting~\eqref{eq:relu_output_mean} equal to the sample mean and~\eqref{eq:relu_output_var} equal to the sample variance measured at the output of the layer, we can solve for $\lambda$ and $\mu$. For the \resnetfifty layer 21 output described earlier, the sample mean over the full validation set is 1.1235656, and the sample variance is 4.9280124, so the numerical solution yields $\lambda = 0.7716595$ and $\mu = -1.4350621$. Now that we have numeric values for these parameters, we can substitute them into~\eqref{eq:relu_output_dist} to obtain a final analytic model for the PDF of data output by layer 21:
\setlength{\maxmin}{\widthof{$-0.144 \leq y < 0$}-\widthof{$y \geq 0$}}
\begin{equation}
     f_{Y}(y) =
     \begin{dcases}
     3.087 e^{4(3.858y+0.554)} & \hspace{\maxmin} y < -0.144 \\
     3.087 e^{-(3.858y+0.554)} & -0.144 \leq y < 0 \\
     0.3087 e^{-(0.3858y+0.554)} & \hspace{\maxmin} y \geq 0 \, .
     \end{dcases}
     \label{eq:relu_output_dist_numeric}
\end{equation}
Fig.~\ref{fig:hist_resnet}(b) shows the fit of this analytic model to the histogram of the feature tensor values output by that layer. We can see that the model captures the salient features of the empirical distribution rather well: sharp peak on the negative side, and slowly decaying exponential on the positive side.  

The feature tensor elements output by the activation function are clipped and quantized using~\eqref{eq:clipquant}. Unlike the quantizer model of~\cite{Banner2019_4bit} for which values are quantized to the midpoint of a quantizer bin, values falling within the first and last bins of our quantizer are quantized to the outer boundaries of those bins. Since these boundaries correspond to the minimum and maximum clipping limits, values that are clipped to ${c}_\mathrm{min}$ or ${c}_\mathrm{max}$ incur no further quantization error. For the $N$-level quantizer of~\eqref{eq:clipquant}, the width of an interior bin is ${\Delta=({c}_\mathrm{max}-{c}_\mathrm{min})/(N-1)}$, 
and the width of each outermost bin is ${\Delta/2}$. Given a PDF $f_{Y}(y)$, the quantization error for values inside the clipping range is
\begin{equation}
    \begin{split}
        &e_{\mathrm{quant}} = 
        \int_{c_\mathrm{min}}^{c_\mathrm{min}+\frac{\Delta}{2}}\!f_Y (y)(y-c_\mathrm{min})^{2}\,dy \\
        &\;\;+
        \sum _{i=1}^{N-2} \left( \int_{{c}_\mathrm{min}+{\frac{\Delta}{2}}
        + \left( i-1 \right) \Delta}^{{c}_\mathrm{min}+{\frac{\Delta}{2}}+i\Delta}
        \!f_Y (y)  \left( y -\left({c}_\mathrm{min} + i\Delta\right) \right) ^{2}
        \,dy \right) \\
        &\;\;+
        \int_{c_\mathrm{max}-\frac{\Delta}{2}}^{c_\mathrm{max}}\!f_Y (y)(y-c_\mathrm{max})^{2}\,dy \, ,
    \end{split}
\end{equation}
where the first and last integrals compute the quantization error for the outermost bins, and the summation portion accumulates quantization error over the interior bins.
The error caused by clipping is
\begin{equation}
    e_{\mathrm{clip}} =
    \int_{-\infty}^{c_\mathrm{min}}\!f_Y (y)(y-c_\mathrm{min})^{2}\,dy
    +
    \int_{c_\mathrm{max}}^{\infty}\!f_Y (y)(y-c_\mathrm{max})^{2}\,dy
    \label{eq:e_clip}
\end{equation}
as was the case in~\cite{Banner2019_4bit}, except that our clipped values incur no further distortion from quantization.

\begin{figure}[tb]
    \centering
    \includegraphics[width=0.3\textwidth,viewport=7.506000 7.794000 722.501978 435.403745,clip]{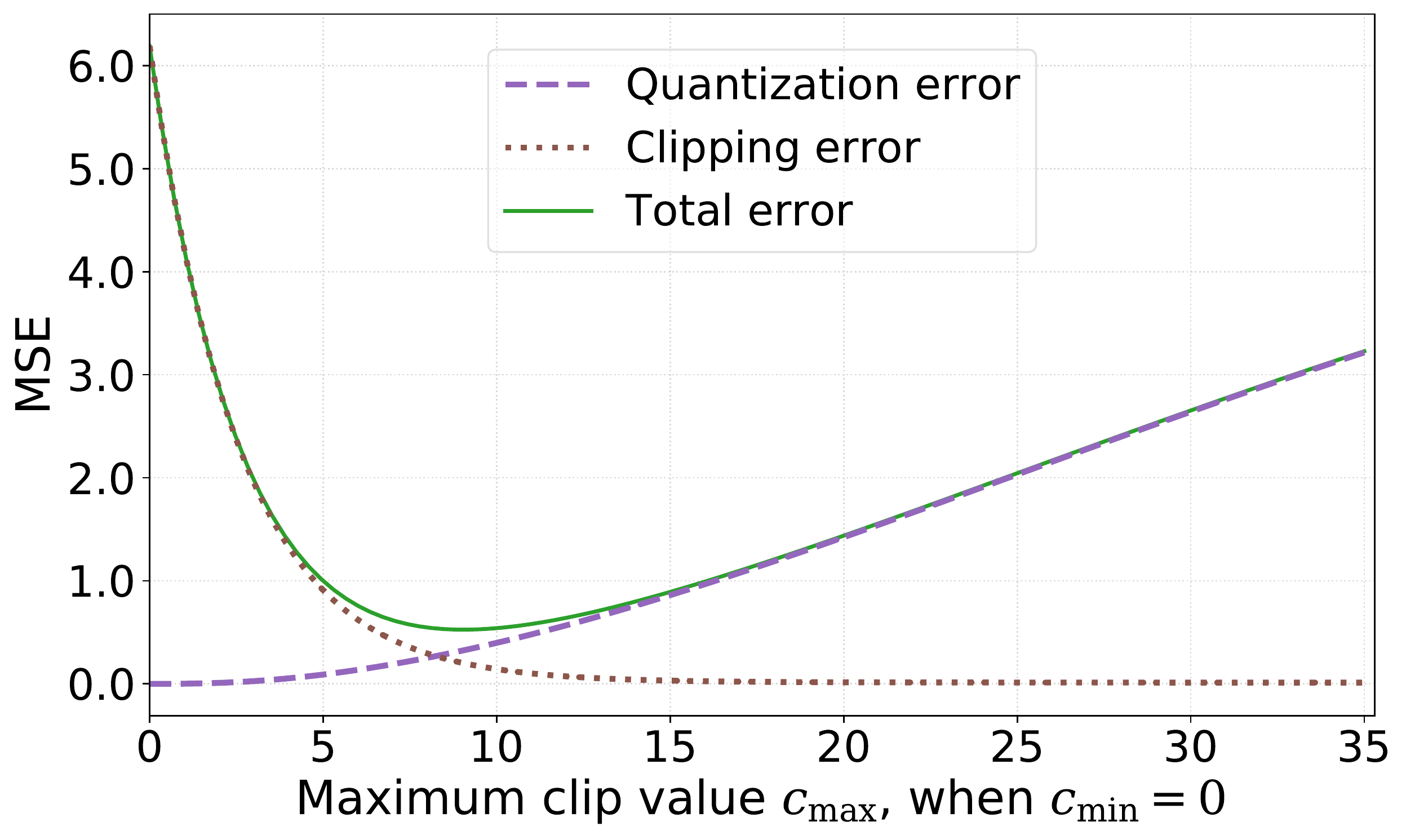}
    \caption{Effects of quantization and clipping error for \resnetfifty analytic model of~\eqref{eq:relu_output_dist_numeric} with clipping and 4-level (2-bit) quantization.}
    \label{fig:error_plot_resnet_addition}
\end{figure}

The total reconstruction error for the clipping and quantizing processes is $e_{\mathrm{tot}} = e_{\mathrm{quant}} + e_{\mathrm{clip}}$.
Given an $N$-level quantizer and the analytic model for the density function $f_{Y}(y)$
from~\eqref{eq:relu_output_dist_numeric},
we can numerically solve for the optimal clipping range $\left[c_\mathrm{min},c_\mathrm{max}\right]$ by minimizing $e_{\mathrm{tot}}$, or for the case when we want $c_\mathrm{min}$ to be zero, we can solve for $c_\mathrm{max}$. An  example illustrating the effects of these errors is shown in Fig.~\ref{fig:error_plot_resnet_addition} for when clipping and quantization with $N=4$ levels is applied to the distribution model in~\eqref{eq:relu_output_dist_numeric}.
As expected, the clipping error $e_{\mathrm{clip}}$ decreases monotonically as $c_\mathrm{max}$ increases, because fewer values are being clipped. We can see from Fig.~\ref{fig:hist_resnet}(b), which is the density plot corresponding to~\eqref{eq:relu_output_dist_numeric}, that feature tensor values $y>5$ are in the tail of the distribution, causing the clipping error curve to level out. Note also from~\eqref{eq:e_clip} that the clipping error is independent of the number of quantizer levels $N$. The quantization error $e_{\mathrm{quant}}$ reduces when $c_\mathrm{max}$ becomes small, given that the width of a quantization bin is also small. Within the clipping ranges of interest, $e_{\mathrm{quant}}$ continues to increase with $c_\mathrm{max}$, however, it will eventually level off because the leftmost bin will become so large that most values are quantized to zero. The total reconstruction error $e_{\mathrm{tot}}$ shown here is the sum of the clipping and quantization error, so we can see that for small clipping ranges, the clipping error dominates, and for large clipping ranges, the quantization error dominates. For this example with $N=4$, the closed-form expression corresponding to $e_{\mathrm{tot}}$ that we want to minimize can be simplified to:
\begin{equation}
\begin{split}
        &e_{\mathrm{tot}} = 6.190 - 0.795 \cdot c_\mathrm{max} \\
        &\cdot \left(e^{\frac{-0.3858}{6} \cdot c_\mathrm{max}} + e^{3(\frac{-0.3858}{6} \cdot c_\mathrm{max})} + e^{5 (\frac{-0.3858}{6} \cdot c_\mathrm{max})} \right) \, .
    \label{eq:e_tot_example}
    \end{split}
\end{equation}

By applying this method over
different values of $N$, we obtain a complete set of closed-form expressions for $e_{\mathrm{tot}}$ that can be minimized to give us clipping ranges that are optimal for an $N$-level quantizer. We can also apply the same technique for modeling the error for YOLOv3 layer 12 activations, whose distribution is
\begin{equation}
     f_{Y}(y) =
     \begin{dcases}
     9.560 e^{4(11.950y+0.369)} & \hspace{\maxmin} y < -0.031 \\
     9.560 e^{-(11.950y+0.369)} & -0.031 \leq y < 0 \\
     0.956 e^{-(1.195y+0.369)} & \hspace{\maxmin} y \geq 0 \, .
     \end{dcases}
     \label{eq:relu_output_dist_numeric_yolo}
\end{equation}
based upon sample mean and variance values of 0.4484323 and 0.5742644, respectively, obtained over the COCO 2017 validation set. We use the same methodology to obtain a model for AlexNet, which uses non-leaky ReLU.

\begin{figure}[tb]
    \begin{minipage}[b]{.9\columnwidth}
        \centering
        \includegraphics[width=\columnwidth,viewport=7.506000 7.794000 721.421978 426.833987,clip]{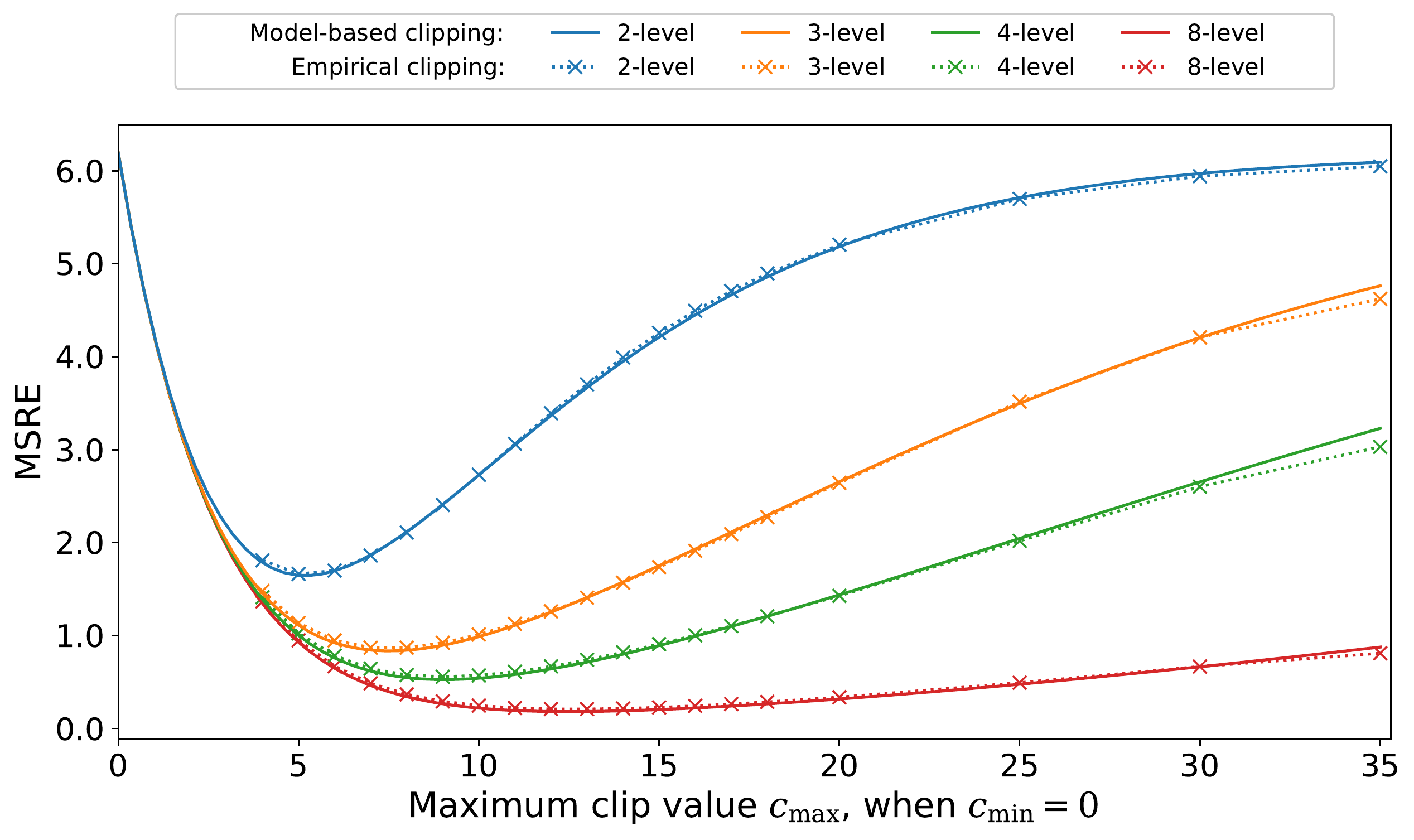}\\
        (a) \resnetfifty
    \end{minipage}
    \vspace{6pt}
    \begin{minipage}[b]{.9\columnwidth}
        \centering
        \includegraphics[width=\columnwidth,viewport=7.506000 7.794000 721.367978 426.833987,clip]{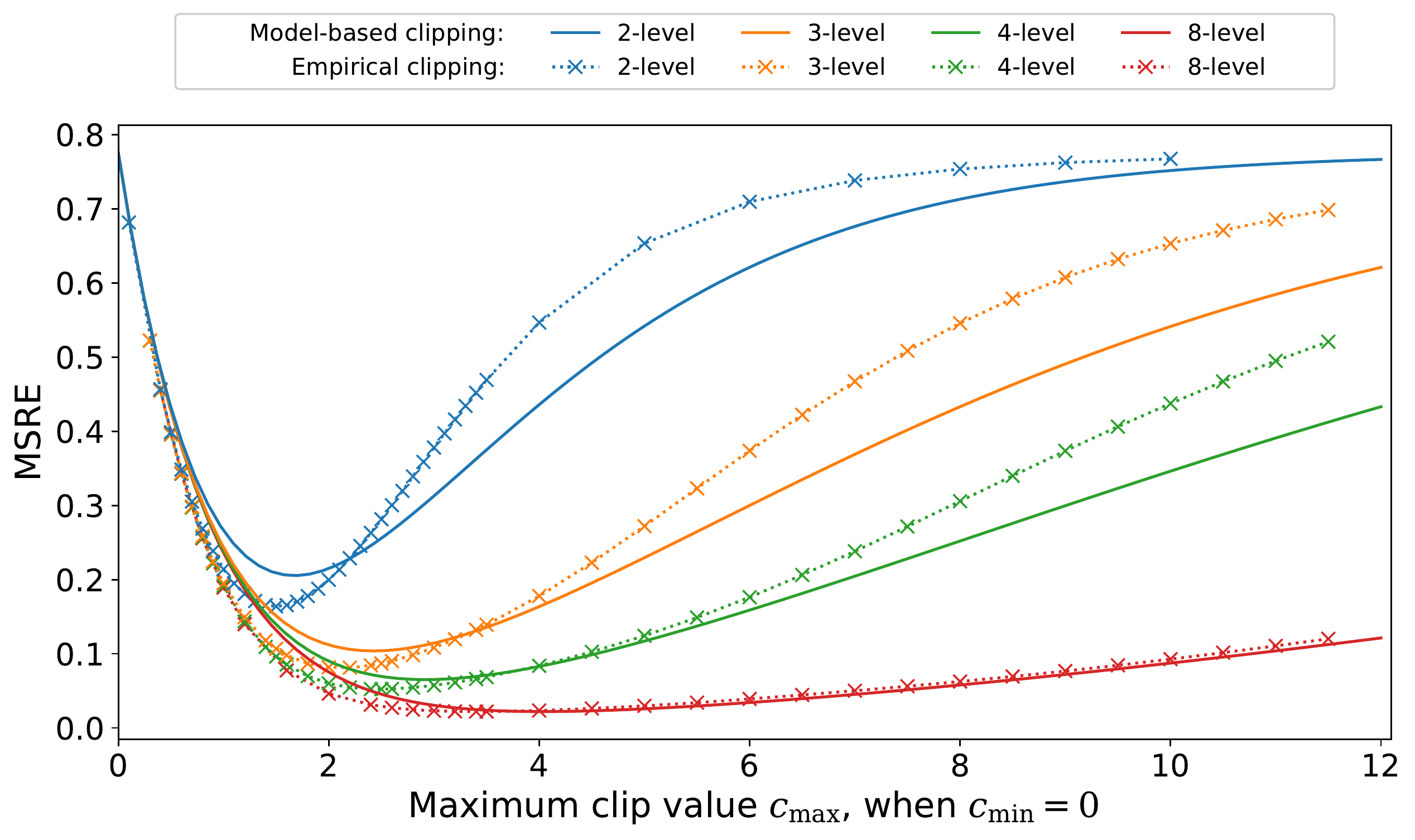}\\
        (b) YOLOv3
    \end{minipage}
    \vspace{6pt}
    \begin{minipage}[b]{.9\columnwidth}
        \centering
        \includegraphics[width=\columnwidth,viewport=7.506000 7.794000 718.793837 426.833987,clip]{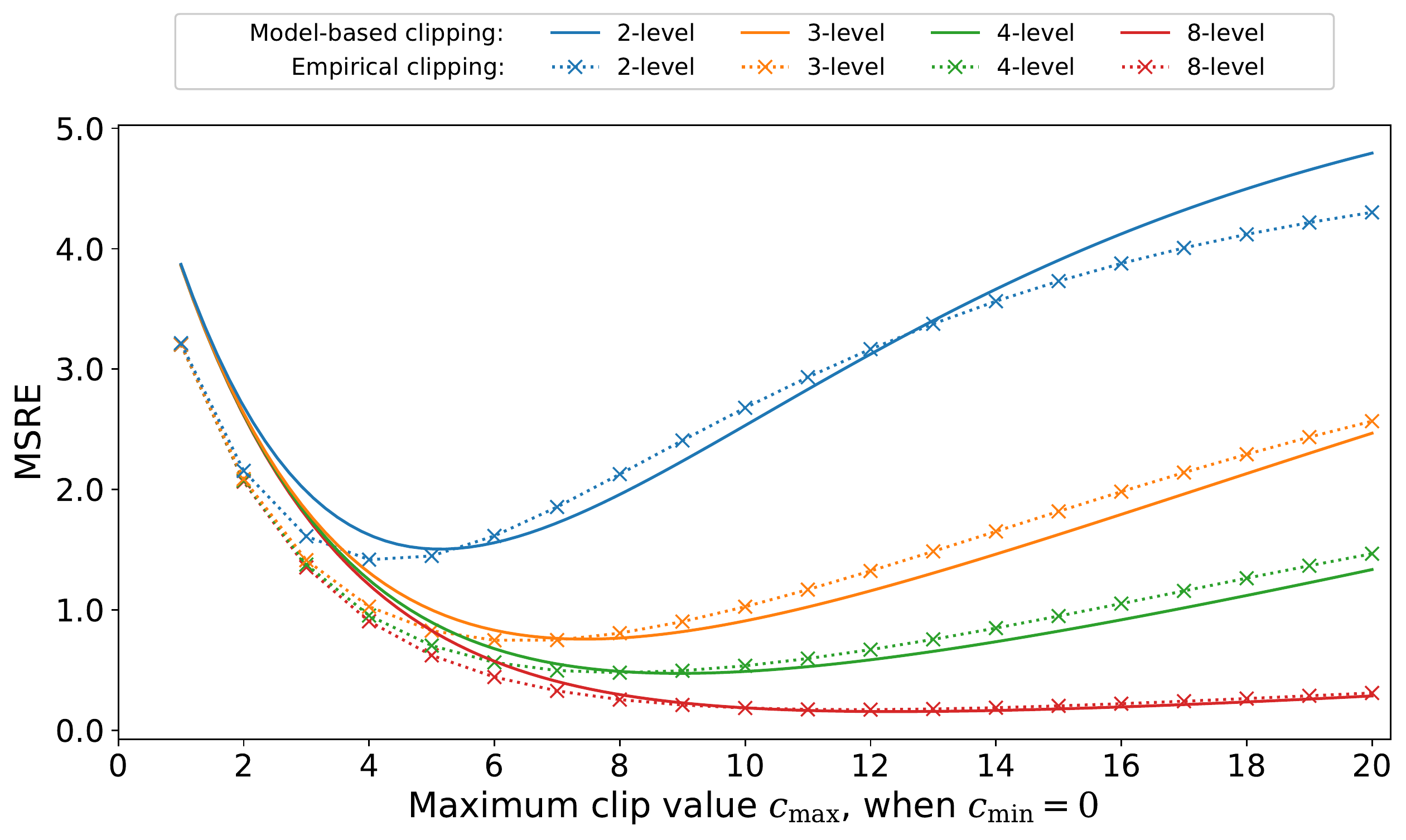}\\
        (c) AlexNet
    \end{minipage}\hfill    
    \caption{Comparison of analytic models for $e_{\mathrm{tot}}$ and the actual error measured experimentally for (a) \resnetfifty layer 21, (b) YOLOv3 layer 12, and (c)~AlexNet layer 4.}
    \label{fig:error_plots}
\end{figure}

We can visualize the accuracy of these models by comparing the analytic $e_{\mathrm{tot}}$ curves to the measured reconstruction error curves we saw earlier in Fig.~\ref{fig:clip_range_resnet_yolo_alexnet}. These comparisons are shown in Fig.~\ref{fig:error_plots}.
For \resnetfifty, the total error computed by model does an excellent job of estimating the actual measured total error for a given clipping range. For YOLOv3 and AlexNet, although the curves do not overlap exactly, their minimum ranges yield corresponding $c_\mathrm{max}$ values that are relatively close to the empirically found optima, which is the intent of these models. The deviation at larger clipping values of the model-based error from the measured error is shown for completeness. This deviation does not negatively impact the overall neural network performance, as we will show in Section~\ref{sec:results_clipping}.

\begin{figure}[tb]
    \begin{minipage}[b]{\columnwidth}
        \centering
        \includegraphics[width=.8\columnwidth,viewport=6.691078 6.696000 1005.052751 72.431998,clip]{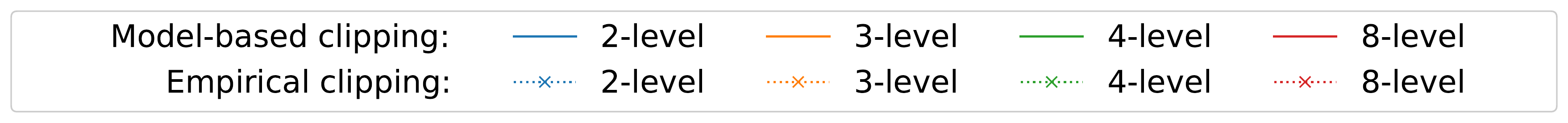}
    \end{minipage}\hfill
    \vspace{1pt}
    \begin{minipage}[b]{0.5\columnwidth}
        \centering
        \includegraphics[width=.95\columnwidth,viewport=7.506000 7.794000 718.793837 616.571981,clip]{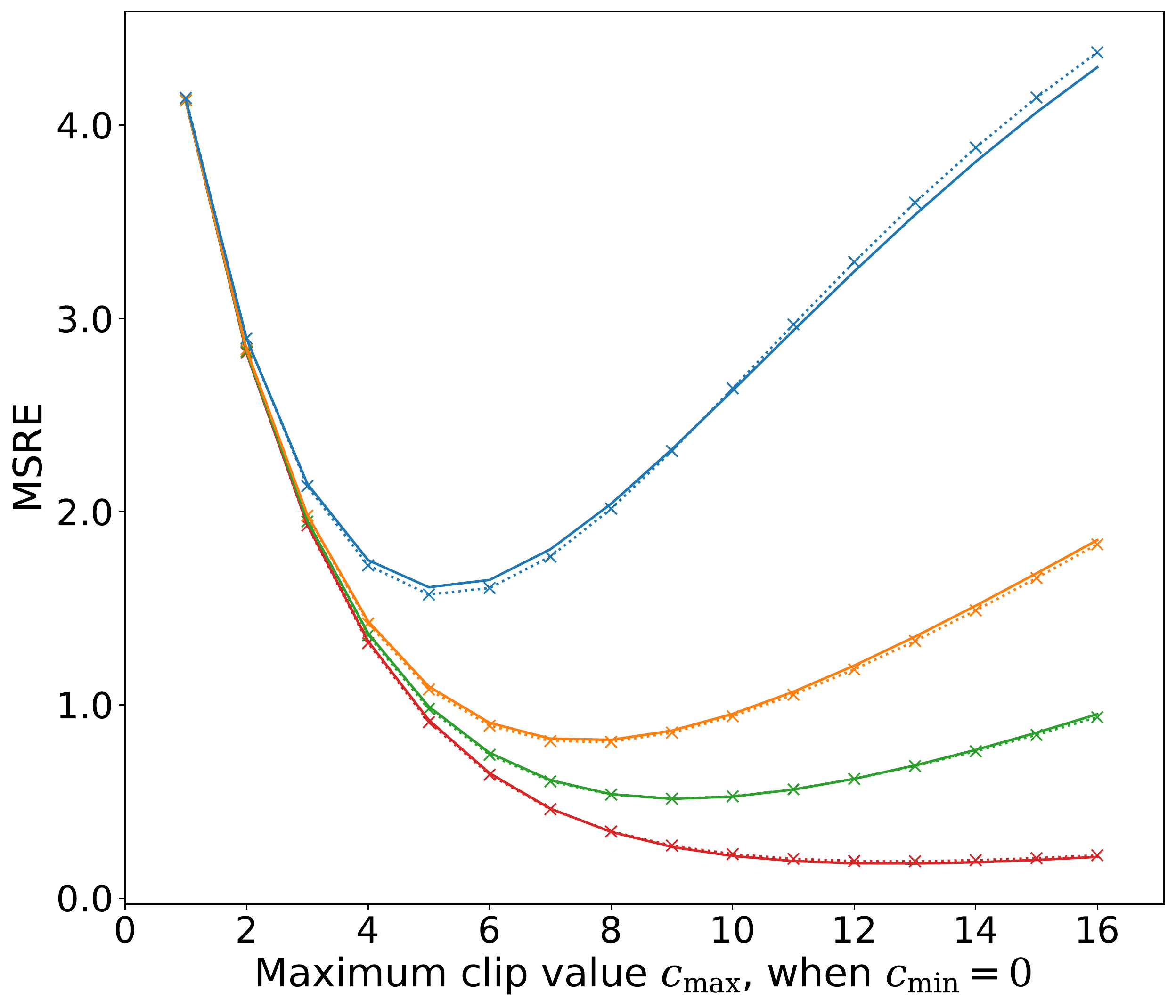}\\
        (a)
    \end{minipage}\hfill
    \begin{minipage}[b]{0.5\columnwidth}
        \centering
        \includegraphics[width=.95\columnwidth,viewport=7.596000 7.992000 712.155002 609.629677,clip]{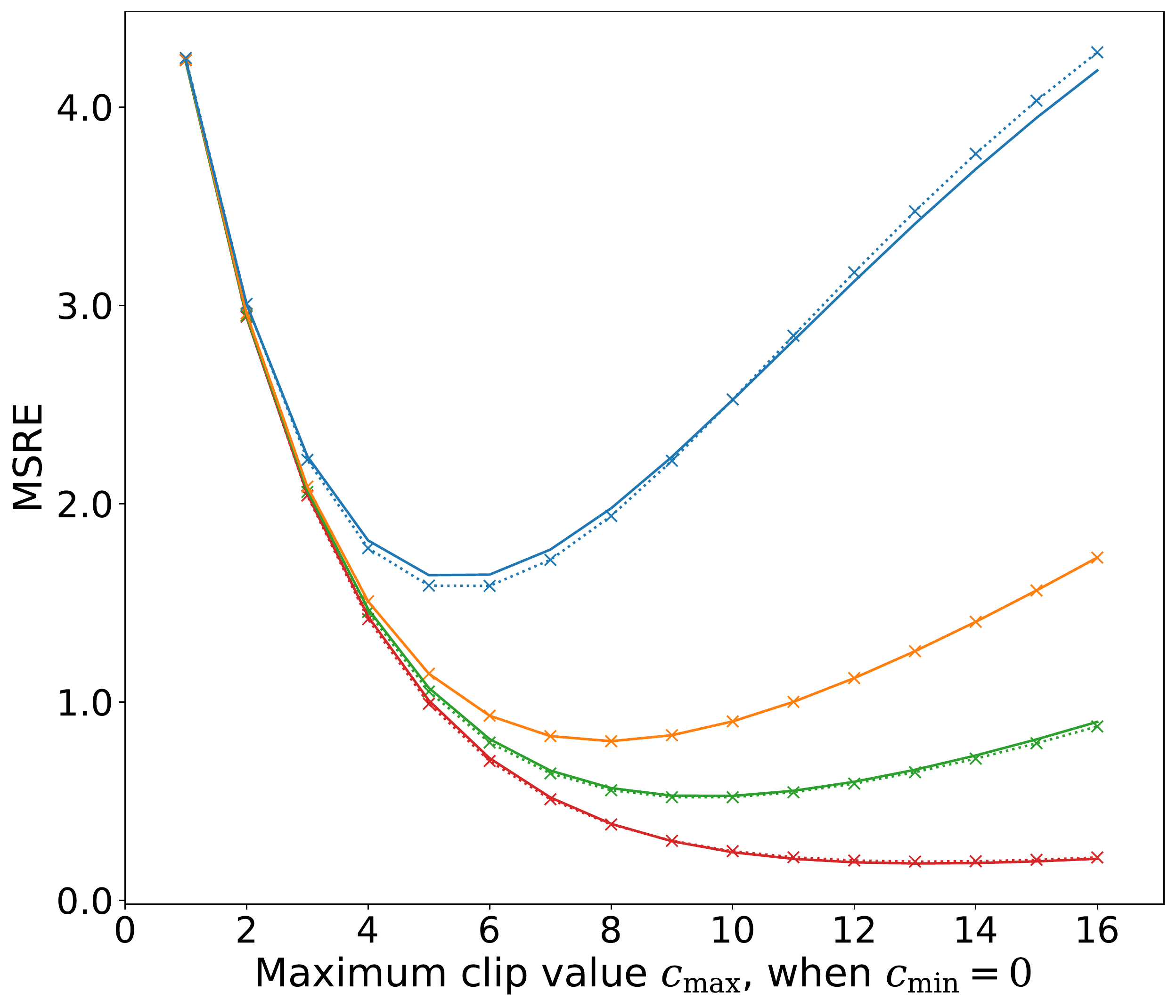}\\
        (b)
    \end{minipage}\hfill    
    \caption{Comparison of analytic models for $e_{\mathrm{tot}}$ and the actual error measured experimentally for (a) \resnetfifty layer 25 and (b) \resnetfifty layer 29.}
    \label{fig:error_plots_resnet_25_29}
\end{figure}

To verify that this model-based method also works well on other layers, for the next two experiments we split \resnetfifty at layers 25 and then at layer 29. These activations correspond to the outputs of the next two shortcut layers after layer 21. Fig.~\ref{fig:error_plots_resnet_25_29} shows that the model provides a good fit to the measured total error for both these layers.

Now that we have a method for optimally clipping and quantizing the feature tensors, we next present methods to improve the compression efficiency of the overall system.

\subsection{Modified entropy-constrained quantization for clipped activations}
A uniform quantizer is  optimal only for signals that are uniformly distributed. Since neural network activations are not uniformly distributed, as seen in the previous section, we need to look at methods for non-uniform quantizer design.
Specifically, we consider entropy-constrained quantization~\cite{Chou1989} in conjunction with novel adjustments designed to improve network accuracy under such quantization.
In conventional quantization, the reconstruction value for each bin of an $\ell^2$-norm optimized non-uniform quantizer corresponds to the centroid (conditional mean) of the data quantized to that bin. Clipped values would therefore be further quantized to the centroid of the outermost quantizer bins. This would, in turn, cause the reconstructed inverse-quantized values to span a range smaller than the optimal clipping range. We showed earlier that with coarse quantization, the accuracy of a DNN can be quite sensitive to the clipping range of a layer's activation. To address this problem, we present a modified entropy-constrained quantizer design process where the reconstruction values of the outermost bins are pinned to ${c}_\mathrm{min}$ and ${c}_\mathrm{max}$, to ensure that the reconstructed activations span the full clipping range. This pinning is only used in outermost bins; the reconstruction values for other bins, as well as bin boundaries,
are not pinned and are optimized by
the algorithm.

Following the notation of~\cite{Girod_ecquant}, the proposed modified entropy-constrained quantizer design process for an $N$-level quantizer is presented in Algorithm~\ref{alg:ec_mod}, with shaded areas indicating how it differs from the conventional entropy-constrained quantizer design approach.
The main modifications are related to pinning the reconstruction values for the boundary bins and using codeword lengths instead of probabilities for computing rate-related terms. The boundary pinning occurs in Step~\ref{stp:ec_mod_update}. Here, the smallest and the largest reconstruction values are always set to the minimum and maximum activation clipping values ${c}_\mathrm{min}$ and ${c}_\mathrm{max}$, respectively. The interior levels are computed as is done in the conventional algorithm.
For computing the rate terms, we replace the probability-based rate estimate,
$\log_{2}(p_n)$, with the known length $b_n$ of the binarized codeword used to represent the bin.

With these modifications in mind, we can now summarize the new quantizer design process. The feature tensor elements output by a split DNN are first clipped in Step 1 to be within $\left[{c}_{\mathrm{min}}, {c}_{\mathrm{max}}\right]$. Next, the values to which input data are quantized, $\hat{x}_n$, are initialized uniformly over the $\left[{c}_{\mathrm{min}}, {c}_{\mathrm{max}}\right]$ interval. In Step 3, the samples used to train the quantizer, which is unrelated to the concept of training weights of a DNN, are assigned to a reconstruction value based on a cost function that uses a Lagrange multiplier $\lambda$. For our experiments with this process, we use feature tensors generated by 100 images from the validation set. For small $\lambda$, the objective is to minimize the quantization error, at the expense of having a larger bit-stream. Conversely, a large $\lambda$ tries to minimize the bit-stream size while allowing the distortion to be large. Thus, $\lambda$ can be used to determine our operating point on a rate-distortion curve.

Once all training samples have been assigned to a reconstruction value, Step 4 updates the reconstruction values by setting them to the average value of the training samples that were assigned to it, i.e. their centroid. However, our modifications pin the first and last reconstruction values to be equal to ${c}_{\mathrm{min}}$ and ${c}_{\mathrm{max}}$, respectively, to ensure that the decoded and reconstructed feature tensors span the optimal $\left[{c}_{\mathrm{min}}, {c}_{\mathrm{max}}\right]$ range. Steps 3 and 4 are repeated until the decrease in the cost function is less than a threshold, or until a certain number of iterations have occurred. Finally, Step 6 computes the decision boundaries between these reconstruction values, also using a Lagrangian cost function. When the quantizer is deployed, each input value is associated with a reconstructed value based on where it falls in relation to the decision thresholds, and the index correspond to the associated reconstruction value is output by the quantizer. Next, we discuss the binarization of this index and subsequent entropy coding.

\begin{algorithm}[!htb]
\small
\caption{Modified entropy-constrained quantizer design process for clipped activations.}
\label{alg:ec_mod}
    \hspace*{\algorithmicindent} \textbf{Input:} Training samples \\
    \hspace*{2cm} $x
    \in \{x_m; m = 0,1,\dotsc,M-1\}$ \\
    \hspace*{1.45cm} Number of quantizer bins $N$ \vspace{1pt} \\
    \tcbox[colback=mypeach,boxrule=0pt,frame hidden,left=0pt, right=0pt, top=0pt, bottom=0pt, left skip=1.42cm,nobeforeafter,after skip=2pt]{Codeword lengths $b_{n}, n \in \{1, \dotsc ,N-1\}$}
    \hspace*{1.45cm} Lagrange multiplier $\lambda$ \vspace{1pt}\\
    \tcbox[colback=mypeach,boxrule=0pt,frame hidden,left=0pt, right=0pt, top=0pt, bottom=0pt, left skip=1.42cm,nobeforeafter]{Activation clipping range $\left[{c}_{\mathrm{min}}, {c}_{\mathrm{max}}\right]$}\\
    \hspace*{\algorithmicindent} \textbf{Output:} Quantizer reconstruction values \\
    \hspace*{2.2cm} $\hat{x}_{n}, n \in \{0,1, \dotsc ,N-1\}$ \\
    \hspace*{1.7cm} Quantizer decision thresholds \\
    \hspace*{2.2cm} $t_{n}, n \in \{1, \dotsc ,N-1\}$
    \vspace{2pt}
\begin{algorithmic}[1]
\State \begin{tcolorbox}[colback=mypeach,boxrule=0pt,frame hidden,left=2pt, right=2pt, top=0pt, bottom=0pt,left skip=-2pt,nobeforeafter,before skip=-10pt]\hspace{-3pt}Clip (clamp) the training samples $x$ to be within $\left[{c}_{\mathrm{min}}, {c}_{\mathrm{max}}\right]$, which is the clipping range applied to the activations\end{tcolorbox}
\State  Initialize the reconstruction values $\hat{x}_n$ for each bin, for $n \in \{0,1,\dotsc,N-1\}$ (e.g., uniform)
\State \label{stp:ec_mod_assign}Assign each training sample $x_m$, $m \in \{0,1,\dotsc,M-1\}$ to quantizer bin $n$ having representative value $\hat{x}_n$ such that the Lagrangian rate-distortion cost is minimized:
    \begin{equation*}
        \argmin_{n} \left[ (x_m - \hat{x}_n)^{2} - \lambda \tcboxmath[colback=mypeach,boxrule=0pt,frame hidden,left=0pt, right=0pt, top=0pt, bottom=0pt]{b_n} \right]
    \end{equation*}
    The subset of samples $x_m$ assigned to bin $n$ is denoted as $\bm{B}_n$.
\State \label{stp:ec_mod_update}Recompute the reconstruction values for each bin:
\parbox{2.3cm}{
   \vspace{1pt}
   \begin{tcolorbox}[colback=mypeach,boxrule=0pt,frame hidden,left=2pt, right=2pt, top=-4pt, bottom=0pt, ams align*]
   \hat{x}_0 &= {c}_{\mathrm{min}} \\
   \hat{x}_{N-1} &= {c}_{\mathrm{max}}
\end{tcolorbox}
   }
\vspace{-4pt}
   \begin{equation*}
   \hspace{-9pt} \hat{x}_n = \frac{1}{|\bm{B}_n|}\sum_{x \in \bm{B}_n} x \, , \quad n \in \{1, \dotsc ,N-2\}, N>2
   \end{equation*}
     where $|\bm{B}_n|$ is the number of samples assigned to bin $n$.
\State Based on the recomputed reconstruction values, recompute the Lagrangian cost function, and repeat Steps~\ref{stp:ec_mod_assign} and~\ref{stp:ec_mod_update} until the reduction in the cost function is less than a threshold.
\State Compute $N-1$ quantizer decision thresholds:
\begin{equation*}
  \begin{split}
    t_{n} = \frac{\hat{x}_n + \hat{x}_{n-1}}{2}
    + \lambda \frac{
    \tcboxmath[colback=mypeach,boxrule=0pt,frame hidden,left=2pt, right=2pt, top=0pt, bottom=0pt]{b_{n} - b_{n-1}}
    }{2(\hat{x}_n - \hat{x}_{n-1})} \, , \\
    n \in \{1, \dotsc ,N-1\}
  \end{split}
\end{equation*}
\end{algorithmic}
\end{algorithm}
\subsection{Binarization and entropy coding}
After quantizing an activation element, an index associated with the selected reconstruction value is coded and signaled to a bit-stream. For the
DNNs
considered in this work, the activation values tend to be
concentrated around zero, as illustrated
in Fig.~\ref{fig:hist_resnet}(b) for the unclipped layer 21 activations of \resnetfifty.
Since we want to achieve good performance when quantizing to very few bins, a truncated unary binarization scheme~\cite{Marpe2003_CABAC} is well suited for this purpose. Given a non-negative integer $n$, this binarization maps $n$ to a binary string comprising $n$ ones followed by a zero, except for the maximum value of $n$ which just maps to $n$ ones. For example, the binarization scheme for a 2-bit (4-level) value maps $n = \{0,1,2,3\} $ to $\{0,10,110,111\}$.

Bit strings produced by binarization can be further compressed using a binary entropy codec.
In this work, we use a simplified version of the Context-based Adaptive Binary Arithmetic Coding (CABAC)~\cite{Marpe2003_CABAC} used in
HEVC and related codecs. One context is used for each bit position in the binarized string. For the 2-bit example described above, three contexts would be used. CABAC builds a separate probability model for each context. When encoding a particular symbol, the probability model associated with the context of that symbol will be used for producing the compressed output following arithmetic coding operations. 

\subsection{Computational complexity}
\label{sec:complexity}
In this section we look at the complexity of the lightweight compression codec and compare it to that of HEVC. When deployed, the lightweight codec itself comprises four steps: clipping, quantization, binarization, and entropy coding. For each feature tensor to be compressed, the clipping step performs two in-place comparisons ($c_{\mathrm{min}}$, $c_{\mathrm{max}}$) for each tensor element. The clipped values are next quantized using~\eqref{eq:clipquant}, whose complexity is equivalent to one addition, two multiplications, and one rounding operation, assuming that the constant values are precomputed. The binarization of the quantizer index can be implemented simply via a lookup table. Given that the lightweight codec typically uses only a few quantization levels, this binarization could even be implemented using a few Boolean logic equations. To entropy code the binarized strings, we use the same entropy coder that HEVC uses (CABAC), except with only a few contexts. To further reduce latency and complexity, some of these operations could be fused into the layer whose output we are compressing.

The complexity of HEVC is broken down by class in~\cite[Table III]{Bossen1012_hevc_complexity}. If we compare the building blocks of our lightweight codec with those listed for HEVC All-Intra (AI), we can get an idea of the relative complexity between these two codecs. Our quantization process would fall into the \textit{TComTrQuant} class. Note that we are only performing quantization, whereas HEVC performs transforms, quantization, and rate-distortion optimization, so our quantization is only a small fraction, perhaps a percent or two, out of the 24.4\% listed in the table. Our binarization and entropy coding operations fall under the entropy-coding classes \textit{TEncSbac}, \textit{TEncEntropy}, and \textit{TEncBinCABAC}. As noted in~\cite{Bossen1012_hevc_complexity}, those classes include scanning and context derivation, whereas our context is simply based on the position of the bit in the binarization table. The lightweight codec's binarization and entropy coding therefore consume a small portion of the 11.8\% total listed for these entropy coding classes. If we estimate that we use few percent from each of these four classes, which is likely a generous overestimate, we can see that our total distribution is less than 10\%. Thus, the lightweight codec is certainly well over 90\% less complex than HEVC. Of course, the actual computation cost depends upon the implementation, but since the modules of the lightweight codec are a subset of those of HEVC, the same optimizations that can be applied to those modules in HEVC can be applied here.

During operation, the lightweight codec needs to know what clipping values $\{c_{\mathrm{min}}$, $c_{\mathrm{max}}\}$ to use just before quantization. To obtain the mean and variance estimates, we used in-line computations on the feature tensor elements at the split layer, over the validation set, before running the codec. For the ImageNet and COCO data sets used in our experiments, the validation sets contained 50k and approximately 5k images, respectively. However, we found that the estimated mean and variance need only a few hundred images to converge; no more than 1k. Since these estimates are computed on unquantized feature tensors, they can easily be computed ahead of time, even during training. This codec is also amenable to adaptive operation if inference is performed in real time while processing video on an edge device. In that case, the measured statistics can adjust based on the most recent few hundred frames.


\section{Experimental results}
\label{sec:experiments}
We applied our lightweight compression technique to activations output from the split layer of a DNN, for two different inference scenarios: \resnetfifty image classification at layer 21, and YOLOv3 object detection at layer 12. The dimensions of the activations at these layers were 32$\times$32$\times$512 and 52$\times$52$\times$256, respectively. Pre-trained network weights were obtained from~\cite{darknet_weights}. The software used to run the experiments was a modified version of the \textit{Darknet} software from~\cite{AlexeyAB_darknet}. For \resnetfifty with network input size 256$\times$256, classification accuracy metrics were obtained directly from the \textit{Darknet} software using the full ImageNet ILSVRC2012 validation data set, which has 50k images. For YOLOv3 with a network input size of 416$\times$416, mAP (IoU~=~0.5) results were obtained using the COCO API~\cite{COCO_API} and the COCO 2017 validation data set, which includes just under 5k images. For experiments using entropy-constrained quantization, the quantizer design algorithms were run on activations output when running the first part of the network on 100 images from the data set. After clipping, quantization, and coding to a bit-stream, the activations were decoded and inverse quantized and then passed to the remainder of the neural network. The bit-streams also included side information needed by the decoder, e.g. $c_{\mathrm{min}}$, $c_{\mathrm{max}}$, $N$, and some dimensional parameters for object detection, which together comprised 24 bytes for object detection and 12 bytes for classification networks.
The size of the compressed data is reported as bits per feature tensor element, i.e., the size of the bit-stream divided by the number of elements in the activation's output feature tensor.

In~\cite{Banner2019_4bit}, Analytical Clipping for Integer Quantization (ACIQ) uses a piecewise linear model for the distribution of the feature tensor values, to approximate the clipping values to be used when all activations of a DNN are quantized to an average of 4 bits. Since ACIQ, like our method, does not require any training to compute the clipping values, we include comparisons to when $c_{\mathrm{max}}$ computed by ACIQ is applied to the output of the layer where our DNN is split. As described in~\cite{Banner2019_4bit}, the $b$ parameter can be estimated from the feature tensor values, assuming that for example the data fits a Laplace density function
$f(x) = \frac{1}{2b}e^{-\frac{|x|}{b}}$. If ReLU is the activation that is applied to this data, then ACIQ assumes $c_{\mathrm{min}}=0$. The equation used to compute $c_{\mathrm{max}}$ can be simplified to:
\begin{equation}
    c_{\mathrm{max}} = b \cdot W(12 \cdot 2^{2M}) \,,
\end{equation}
where $W()$ is the Lambert W function and $M$ is the number of bits to which elements are quantized. Although~\cite{Banner2019_4bit} only quantizes to an integer number of bits, the purpose of our quantization is for subsequent compression, so we can allow for non-integer bit-widths by substituting $M=\log_{2}(N)$ with $N$-level quantization.

\subsection{Clipping and uniform quantization performance}
\label{sec:results_clipping}
\begin{figure}[tb]
    \centering
    \begin{minipage}[b]{.9\columnwidth}
        \centering
        \includegraphics[width=\columnwidth,viewport=7.200000 7.794000 703.143682 421.091987,clip]{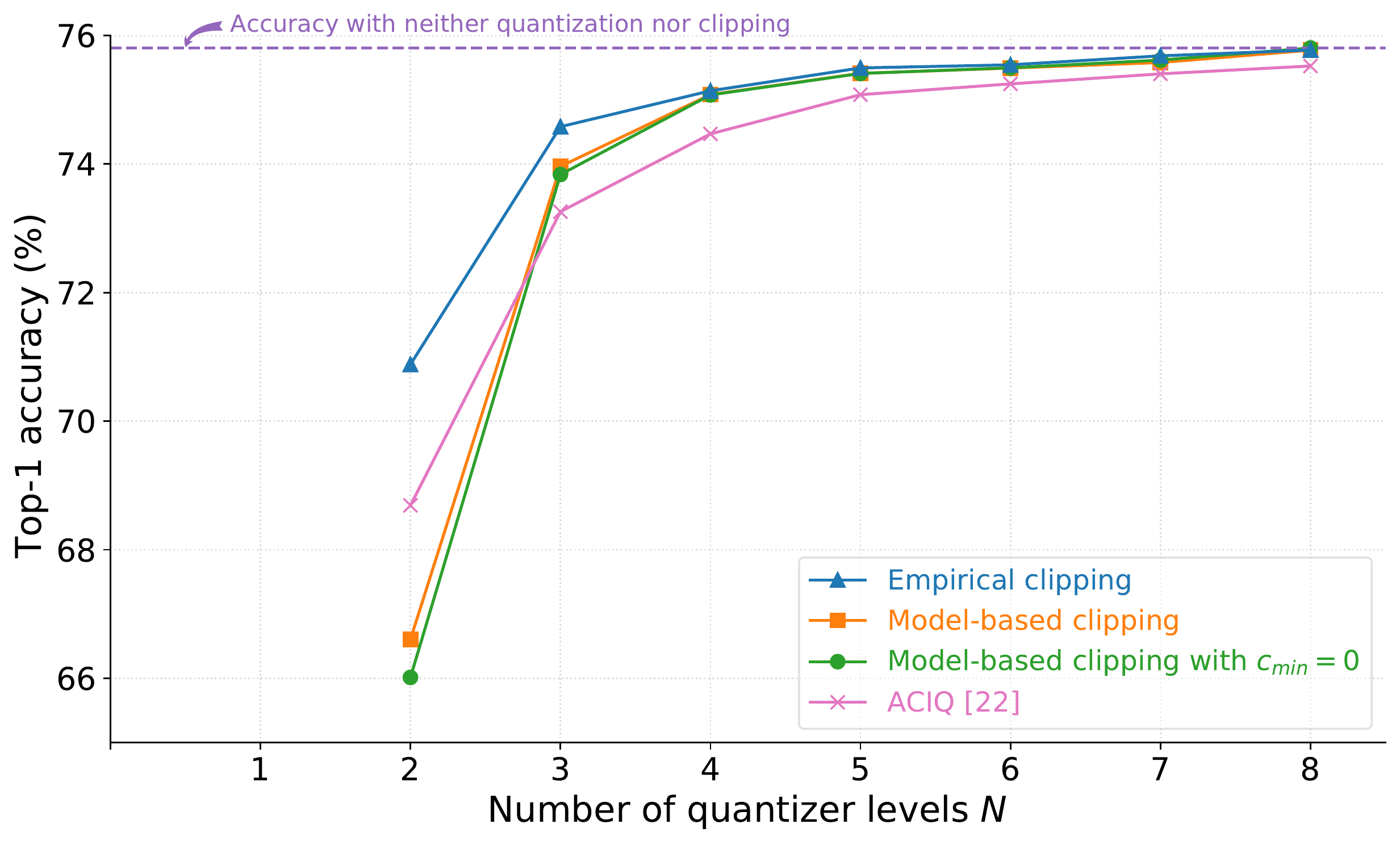}\\
        (a) \resnetfifty
    \end{minipage}
    \vspace{6pt}
    \begin{minipage}[b]{.9\columnwidth}
        \centering
        \includegraphics[width=\columnwidth,viewport=7.200000 7.794000 703.143682 420.965987,clip]{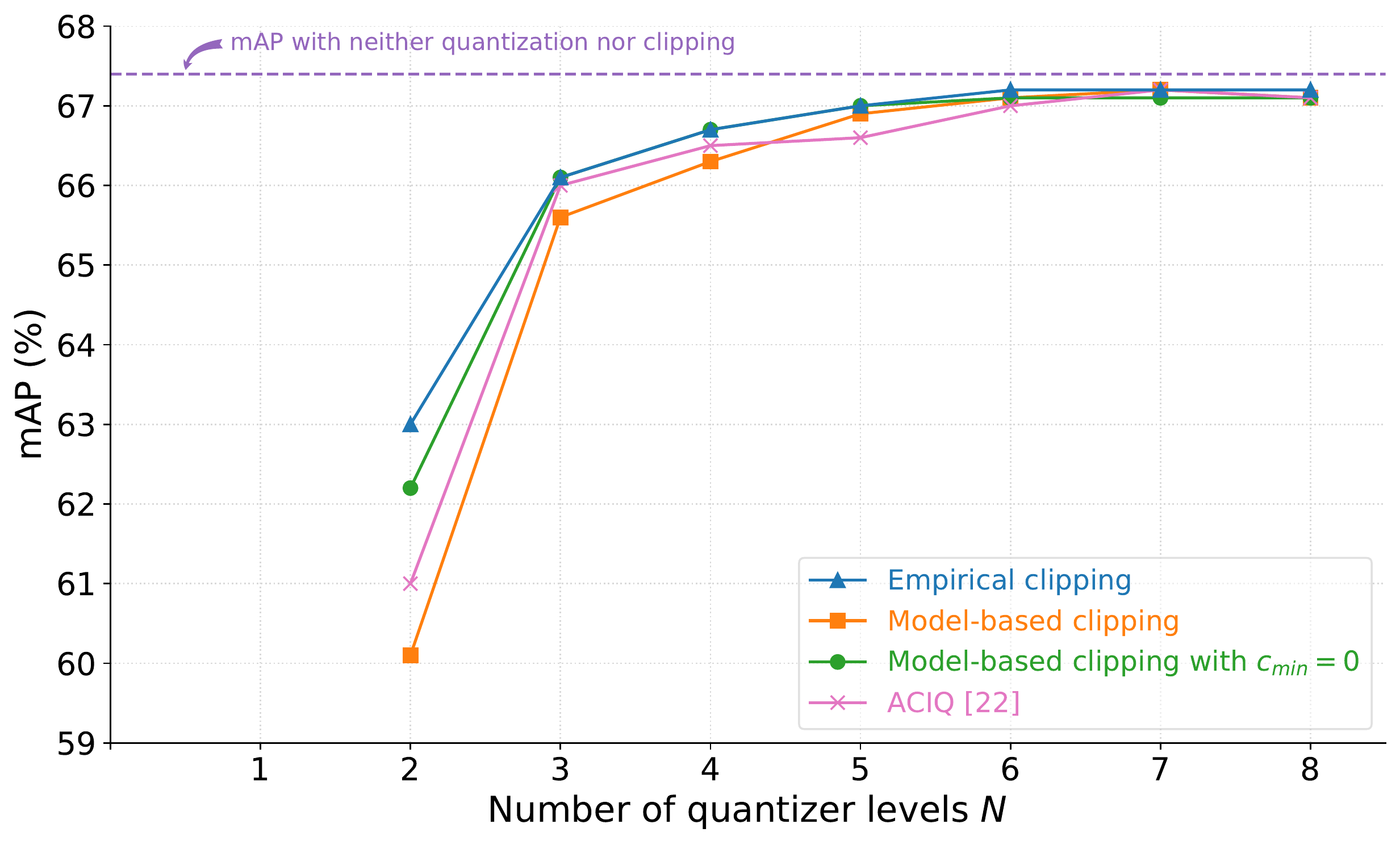}\\
        (b) YOLOv3
    \end{minipage}
    \vspace{6pt}
    \begin{minipage}[b]{.9\columnwidth}
        \centering
        \includegraphics[width=\columnwidth,viewport=7.200000 7.794000 703.143682 420.965987,clip]{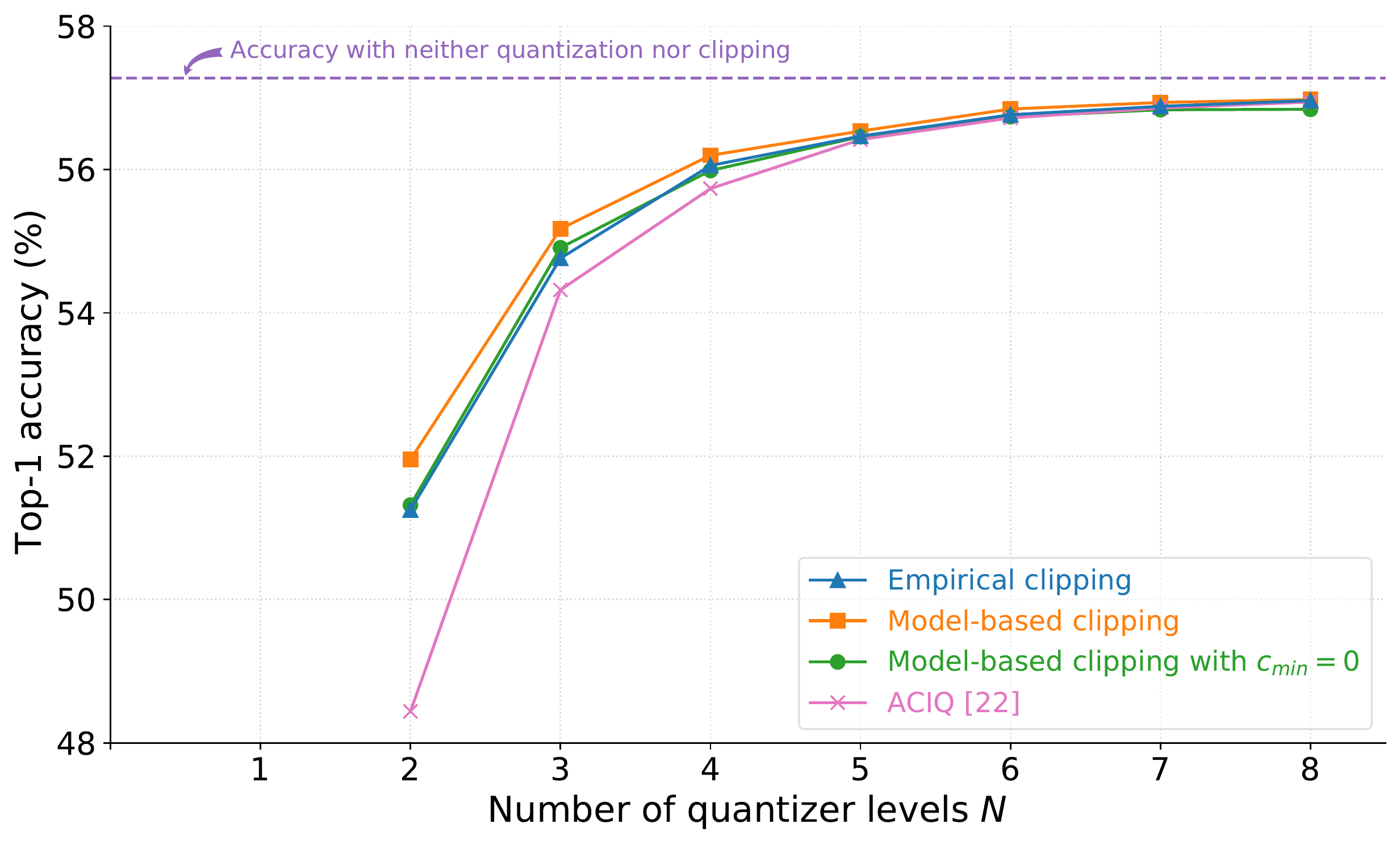}\\
        (c) AlexNet
    \end{minipage}\hfill  
    \caption{Performance of optimal empirical, model-based clipping, and ACIQ on uniform $N$-level quantization activations output by (a) \resnetfifty layer 21, (b) YOLOv3 layer 12, and (c) AlexNet layer 4.}
    \label{fig:resnet_yolo_clip_model}
\end{figure}

Fig.~\ref{fig:resnet_yolo_clip_model}(a) shows the Top-1 performance of \resnetfifty when clipping and $N$-level quantization is applied to the activations output from layer 21. The empirical curve is obtained by running the network over the full validation set with $c_\mathrm{min}=0$ and $c_\mathrm{max}$ set to the value that yields the best accuracy from Fig.~\ref{fig:clip_range_resnet_yolo_alexnet}(a). The curves based on our asymmetric Laplace distribution model are shown both for when the optimal $c_\mathrm{max}$ is obtained after fixing $c_\mathrm{min}=0$, and for when $c_\mathrm{min}$ is not constrained. Plots corresponding to YOLOv3 layer 12 and AlexNet layer 4 are shown in Fig.~\ref{fig:resnet_yolo_clip_model}~(b) and (c), respectively. Note that the performance is plotted vs. the number of quantizer levels $N$ ranging from 2 to 8, which corresponds to between 1 and 3 bits before binarization and entropy coding are applied.
We can see that for 4-level (2-bit) or 5-level and finer quantization, the model with $c_\mathrm{min}$ constrained to zero does a particularly good job of estimating the empirically determined optimal clipping range $c_\mathrm{max}$. For extremely coarse quantization, e.g. 2-level (1-bit) and 3-level quantization, the model's deviation from empirically determined values is expected, given that the model is based on minimizing MSRE and clipping error, and minimizing MSRE does not maximize the overall neural network performance in these cases, as discussed in Section~\ref{sec:clipping}. By removing the constraint $c_\mathrm{min}=0$, the performance can be slightly better or worse than with the constraint, especially with 2-level (1-bit) quantization. The empirical and model-based clipping ranges obtained from these experiments are summarized in Table~\ref{tab:clipping_ranges}, along with the maximum clipping values computed using ACIQ. We can see that applying the constraint $c_\mathrm{min}=0$ has essentially no effect on the size of the clipping interval, i.e. $\left[c_\mathrm{min},c_\mathrm{max}\right]$ is shifted to $\left[0,c_\mathrm{max}-c_\mathrm{min}\right]$. For 4-level (2-bit) and finer quantization, the effects of this constraint on overall network performance are negligible, and therefore fixing $c_\mathrm{min}=0$ may be preferable for ease of implementation. We also can see that as the number of quantization levels decreases, generally the optimal clipping range decreases as well, as discussed in Section~\ref{sec:clipping}.

When using ACIQ, the $c_\mathrm{max}$ values are computed based on a linear approximation to the data's distribution. We can see from Table~\ref{tab:clipping_ranges} that for quantizers having few levels, the $c_\mathrm{max}$ values from ACIQ are generally higher than our empirical and model-based values. As we saw earlier in Fig.~\ref{fig:clip_range_resnet_yolo_alexnet}, the network performance is sensitive to changes in $c_\mathrm{max}$ with coarse quantization. As the quantizer becomes finer, the performance using ACIQ approaches that of the empirical and model-based methods, given that the range of acceptable $c_\mathrm{max}$ becomes wider.
\begin{table}[tb]
    \small
    \centering
    \caption{Empirical and model-based optimal clipping ranges with $N$-level quantization.}
    \label{tab:clipping_ranges}
    \begin{minipage}[b]{\columnwidth}
    \resizebox{\columnwidth}{!}{
    \begin{tabular}{ccrrrrrrrr}
  &  & \multicolumn{4}{c}{\textbf{\resnetfifty}} & \multicolumn{4}{c}{\textbf{YOLOv3}} \\
  &  & \multicolumn{2}{c}{$c_\mathrm{min}$ set to 0} & \multicolumn{2}{c}{$c_\mathrm{min}$ unconstrained} & \multicolumn{2}{c}{$c_\mathrm{min}$ set to 0} & \multicolumn{2}{c}{$c_\mathrm{min}$ unconstrained} \\
\cmidrule[\heavyrulewidth](lr){3-4} \cmidrule[\heavyrulewidth](lr){5-6} \cmidrule[\heavyrulewidth](lr){7-8} \cmidrule[\heavyrulewidth](lr){9-10}
 & & empirical & model & \multicolumn{2}{c}{model} & empirical & model & \multicolumn{2}{c}{model} \\
$N$ & bits & $c_\mathrm{max}$ & $c_\mathrm{max}$ & $c_\mathrm{min}$ & $c_\mathrm{max}$ & $c_\mathrm{max}$ & $c_\mathrm{max}$ & $c_\mathrm{min}$ & $c_\mathrm{max}$ \\
\cmidrule(lr){1-1} \cmidrule(lr){2-2} \cmidrule(lr){3-3} \cmidrule(lr){4-4} \cmidrule(lr){5-6} \cmidrule(lr){7-7} \cmidrule(lr){8-8} \cmidrule(lr){9-10}
2 & 1 &  7.0 &  5.184 &  0.361 &  5.544 & 1.95 & 1.674 &  0.171 & 1.844 \\
3 &   &  9.0 &  7.511 &  0.147 &  7.658 & 2.90 & 2.425 &  0.087 & 2.512 \\
4 & 2 & 10.0 &  9.036 &  0.053 &  9.089 & 3.20 & 2.918 &  0.047 & 2.965 \\
5 &   & 12.0 & 10.175 &  0.001 & 10.176 & 3.20 & 3.285 &  0.026 & 3.311 \\
6 &   & 12.0 & 11.084 & -0.030 & 11.054 & 3.40 & 3.579 &  0.012 & 3.591 \\
7 &   & 12.0 & 11.842 & -0.051 & 11.792 & 4.00 & 3.824 &  0.003 & 3.826 \\
8 & 3 & 14.0 & 12.492 & -0.065 & 12.427 & 4.30 & 4.033 & -0.004 & 4.030 
    \end{tabular}
    }
    \end{minipage}
    \vspace{6pt}
    \begin{minipage}[b]{\columnwidth}
    \centering
    \resizebox{\columnwidth}{!}{
    \begin{tabular}{ccrrrrrrr}
  &  &  \multicolumn{4}{c}{\textbf{AlexNet}} \\
  &  & \multicolumn{2}{c}{$c_\mathrm{min}$ set to 0} & \multicolumn{2}{c}{$c_\mathrm{min}$ unconstrained}  &
  \multicolumn{3}{c}{$\bm{c_\mathrm{max}}$ \textbf{computed using ACIQ~\cite{Banner2019_4bit} }}\\
\cmidrule[\heavyrulewidth](lr){3-4} \cmidrule[\heavyrulewidth](lr){5-6}
 & & empirical & model & \multicolumn{2}{c}{model} \\
$N$ & bits & $c_\mathrm{max}$ & $c_\mathrm{max}$ & $c_\mathrm{min}$ & $c_\mathrm{max}$ & \resnetfifty & YOLOv3 & AlexNet \\
\cmidrule(lr){1-1} \cmidrule(lr){2-2} \cmidrule(lr){3-3} \cmidrule(lr){4-4} \cmidrule(lr){5-6} \cmidrule(lr){7-9}
2 & 1 &  5.0 &  5.134 & 0.415 & 5.549  &  5.722 & 2.460 &  7.728 \\
3 &   &  8.0 &  7.439 & 0.230 & 7.669  &  6.964 & 2.994 &  9.406 \\
4 & 2 &  9.0 &  8.950 & 0.150 & 9.100  &  7.878 & 3.387 & 10.639 \\
5 &   & 11.0 & 10.078 & 0.107 & 10.185 &  8.603 & 3.698 & 11.617 \\
6 &   & 11.0 & 10.979 & 0.081 & 11.060 &  9.203 & 3.957 & 12.428 \\
7 &   & 12.0 & 11.729 & 0.064 & 11.793 &  9.717 & 4.177 & 13.121 \\
8 & 3 & 14.0 & 12.373 & 0.052 & 12.425 & 10.166 & 4.370 & 13.727 \\
    \end{tabular}
    }    
     \end{minipage}\hfill
\end{table}
\begin{figure}[tb]
    \centering
    \begin{minipage}[b]{.9\columnwidth}
        \centering
        \includegraphics[width=\columnwidth,viewport=7.200000 7.218000 701.171979 421.865987,clip]{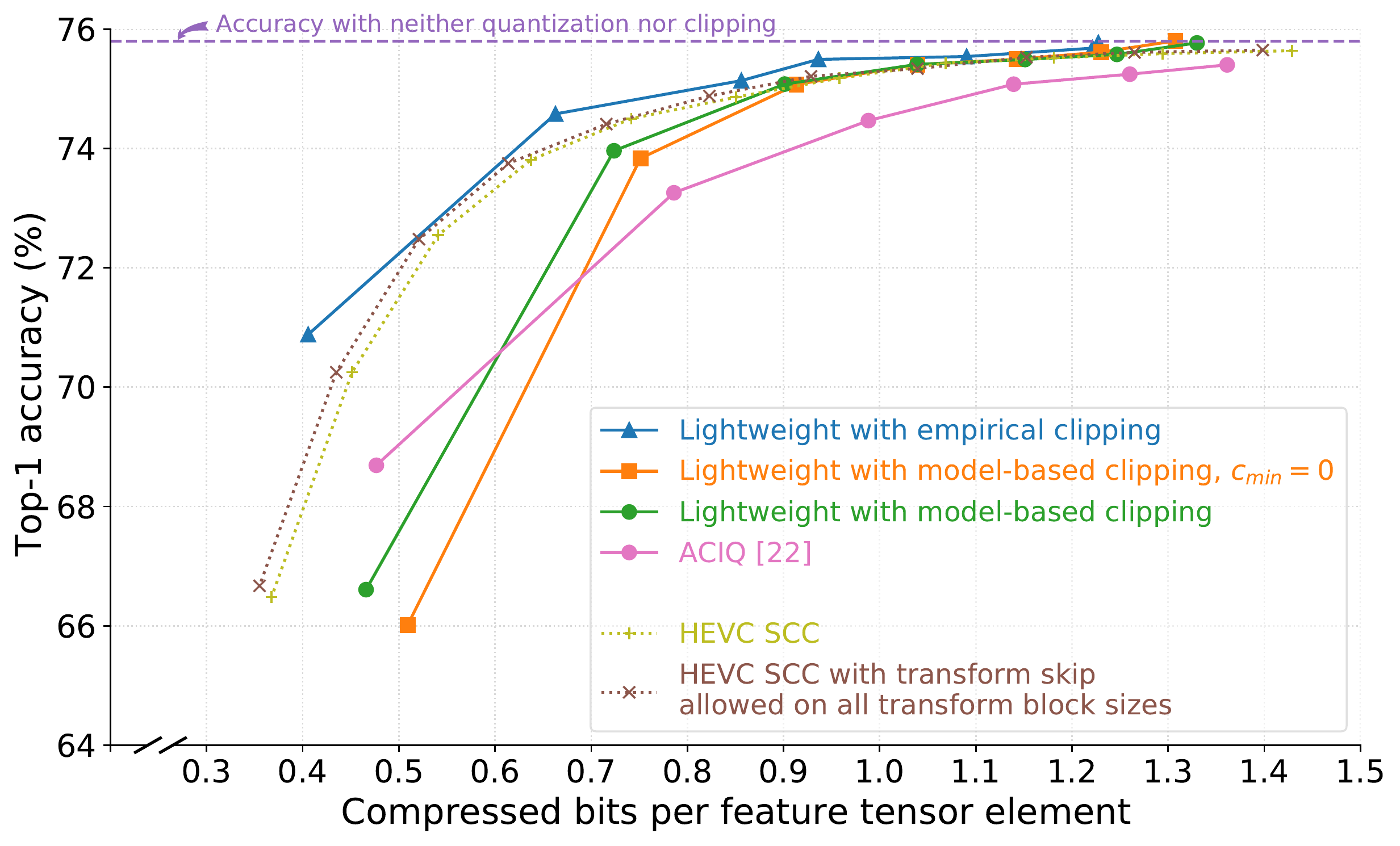}\\
        (a) \resnetfifty
    \end{minipage}
    \vspace{6pt}
    \begin{minipage}[b]{.9\columnwidth}
        \centering
        \includegraphics[width=\columnwidth,viewport=7.200000 7.218000 701.297979 421.829987,clip]{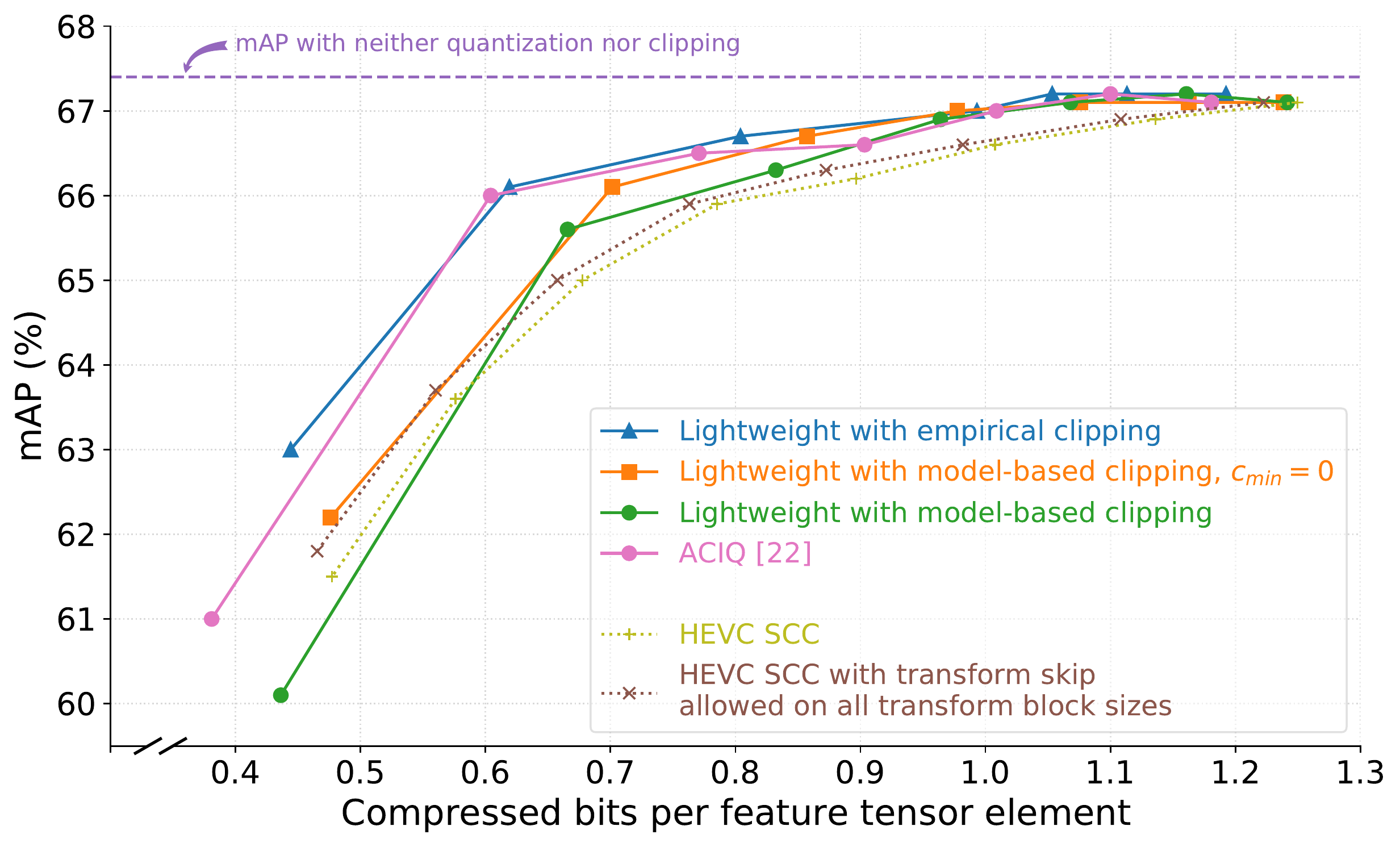}\\
        (b) YOLOv3
    \end{minipage}\hfill    
    \caption{Lightweight compression performance using model-based and empirical clipping with uniform quantization when applied to (a) \resnetfifty layer 21 activations, (b) YOLOv3 layer 12 activations; including comparison to HEVC-SCC. Empirical clipping uses $c_{\mathrm{min}}=0$.}
    \label{fig:resnet_yolo_RD}
\end{figure}

\subsection{Lightweight compression system performance}
\label{sec:results_compression}
Distortion-rate plots showing the overall performance of \resnetfifty and YOLOv3 when lightweight compression is used on layer 21 and layer 12 activations, respectively, are shown in Fig.~\ref{fig:resnet_yolo_RD}(a) and (b), for when $N$-level uniform quantization is used with $N={2,3,...,8}$. The rate or bit-count points in these plots have a one-to-one correspondence to the $N$-level points in Fig.~\ref{fig:resnet_yolo_clip_model}. To compute the amount of compressed bits per feature tensor element, we divide the size of the final output bit-stream by the total number of feature tensor elements that were coded. We also account for the side information included in the header of the bit-streams, which requires 12 bytes for image classification networks and 24 bytes for object-detection networks. This side information includes the original input image dimensions, $c_\mathrm{min}$, $c_\mathrm{max}$, and for object detection the input dimensions of the first layer are signaled so that bounding box coordinates around objects can be computed.

For \resnetfifty, Fig.~\ref{fig:resnet_yolo_RD}(a) shows that the activations could be quantized to 8 levels (3 bits) with no loss relative to when no quantization or clipping was used, and at 4 levels (2 bits), the drop in Top-1 accuracy was well below 1\%. One-bit quantization was feasible with \resnetfifty and YOLOv3, which yielded 4.9\% and 4.8\% losses, respectively, with corresponding compressed sizes of 0.41 and 0.45 bits per element. Additional bit reductions are possible, e.g. to 0.39 bits per element as shown in~\cite{Cohen_quantcode_ICME2020} when nonzero $c_{\mathrm{min}}$ values were used with empirical clipping on YOLOv3. We saw in Fig.~\ref{fig:resnet_yolo_clip_model}(a) for \resnetfifty that the model-based clipping and empirical clipping yielded equivalent performance at 4-level and finer quantization. When the actual compressed rate is used, the distortion-rate performance in Fig.~\ref{fig:resnet_yolo_RD}(a) shows that 6-level (between 2 and 3 bits) quantization and finer yields equivalent performance. For YOLOv3, the model-based and empirical clipping yielded similar performance with 5-level and finer quantization, and for
distortion-rate performance, the model worked well with 3-level and finer quantization. Additionally, for YOLOv3 we were able to use the model to obtain clipping ranges that yielded better performance than those used in the empirical study. For YOLOv3, we showed in~\cite{Cohen_quantcode_ICME2020} that no loss in performance occurred when quantizing all the way down to 16 levels (4 bits), which is why we show here the performance for 8 and fewer levels. The drop in mAP was less than 1\% with a 4-level (2-bit) quantizer.

We can also observe here the effect that $c_\mathrm{max}$ has on the compressed bit-stream size. For example, the leftmost point on the empirical curve in Fig.~\ref{fig:resnet_yolo_RD}(b) corresponds to 2-level quantization with $c_\mathrm{max}$=1.95. With ACIQ, $c_\mathrm{max}$=2.46. We can see in Fig.~\ref{fig:clip_range_resnet_yolo_alexnet} that 2.46 is beyond the optimal clip range for 2-level quantization. However, a wider clip range with uniform quantization causes the quantization bins to become wider. Since the distribution of the data being quantized is denser near zero, more elements will be quantized to the first quantizer bin, whose binarized representation uses only one bit, thus reducing the overall size of the compressed bit-stream.

Fig.~\ref{fig:resnet_yolo_RD} also shows the performance when coding the activations using the HM16.20~\cite{HM16.20} implementation of the HEVC screen content coding extension (HEVC-SCC)~\cite{hevc_std_2017}. HEVC-SCC includes tools that help with the coding of non-camera-captured pictures. As shown in~\cite{Choi2018NearLosslessDF}, activation channels arranged as pictures exhibit much high-frequency content. HEVC-SCC includes a transform skip (TS) mode that is available for all transform block sizes, so we show results when enabling TS for 4$\times$4 blocks only, and for when enabled for all block sizes. Each set of activation channels were quantized to 8 bits and mosaicked into an 832$\times$832 picture for YOLOv3 and to 1024$\times$512 for \resnetfifty. Given the fineness of the quantizer, clipping was not necessary. The mosaicked feature tensors for the validation set were coded by HEVC-SCC as an all-Intra sequence of monochrome (4:0:0) 8-bit pictures.
Even with the improved performance with TS on all block sizes, the lightweight compression system outperformed HEVC-SCC by up to 1.3\%, depending upon rate.

\begin{figure}[tb]
    \centering
    \includegraphics[width=0.42\textwidth,viewport=7.200000 7.218000 701.063979 421.811987,clip]{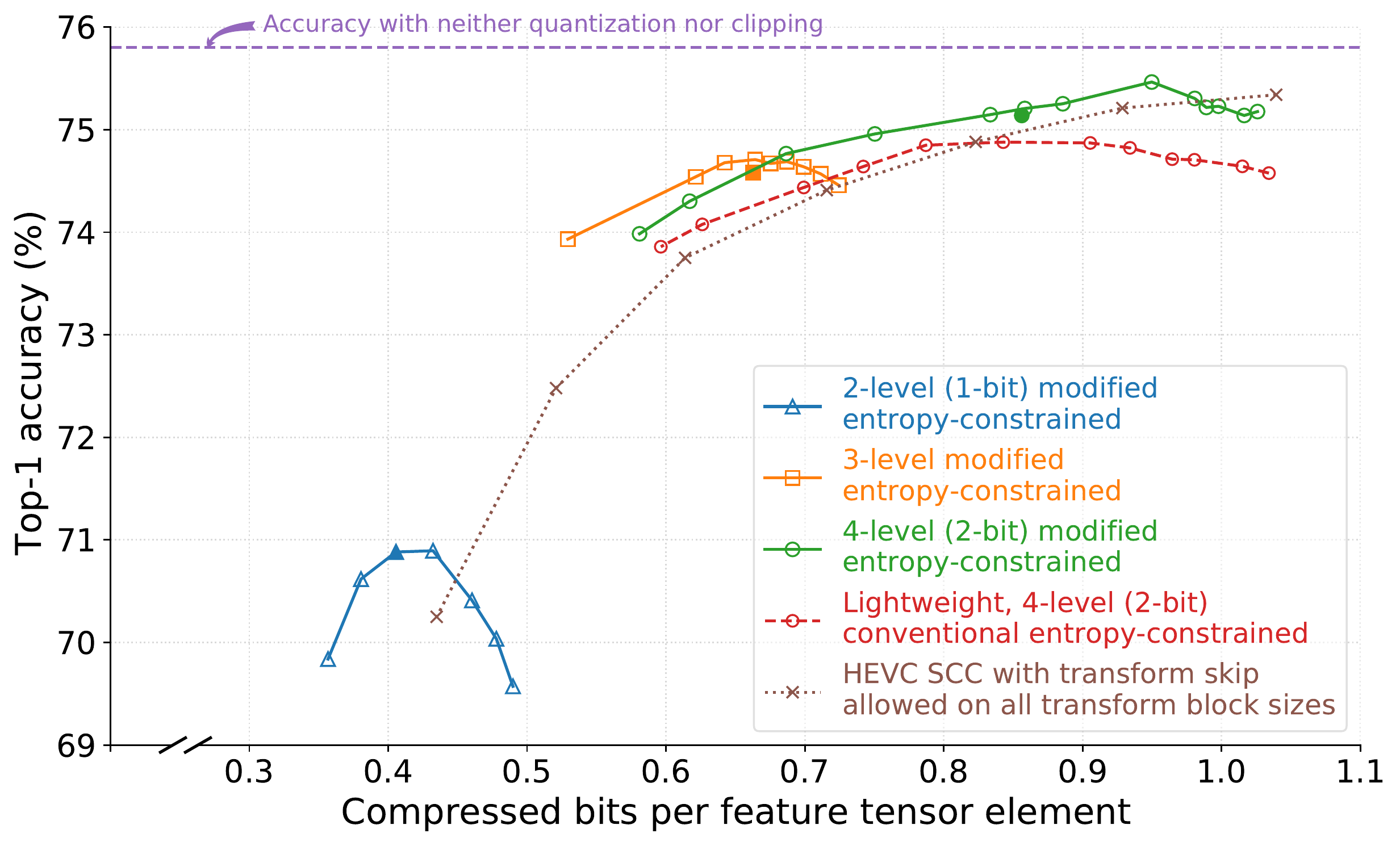}
    \caption{Top-1 accuracy for \resnetfifty using lightweight compression at layer 21 with modified entropy-constrained quantization. Filled markers indicate corresponding performance using uniform quantization.}
    \label{fig:resnet50_top1_plot}
\end{figure}

\begin{figure}[tb]
    \centering
    \includegraphics[width=0.42\textwidth,viewport= 7.200000 7.218000 701.099979 421.829987,clip]{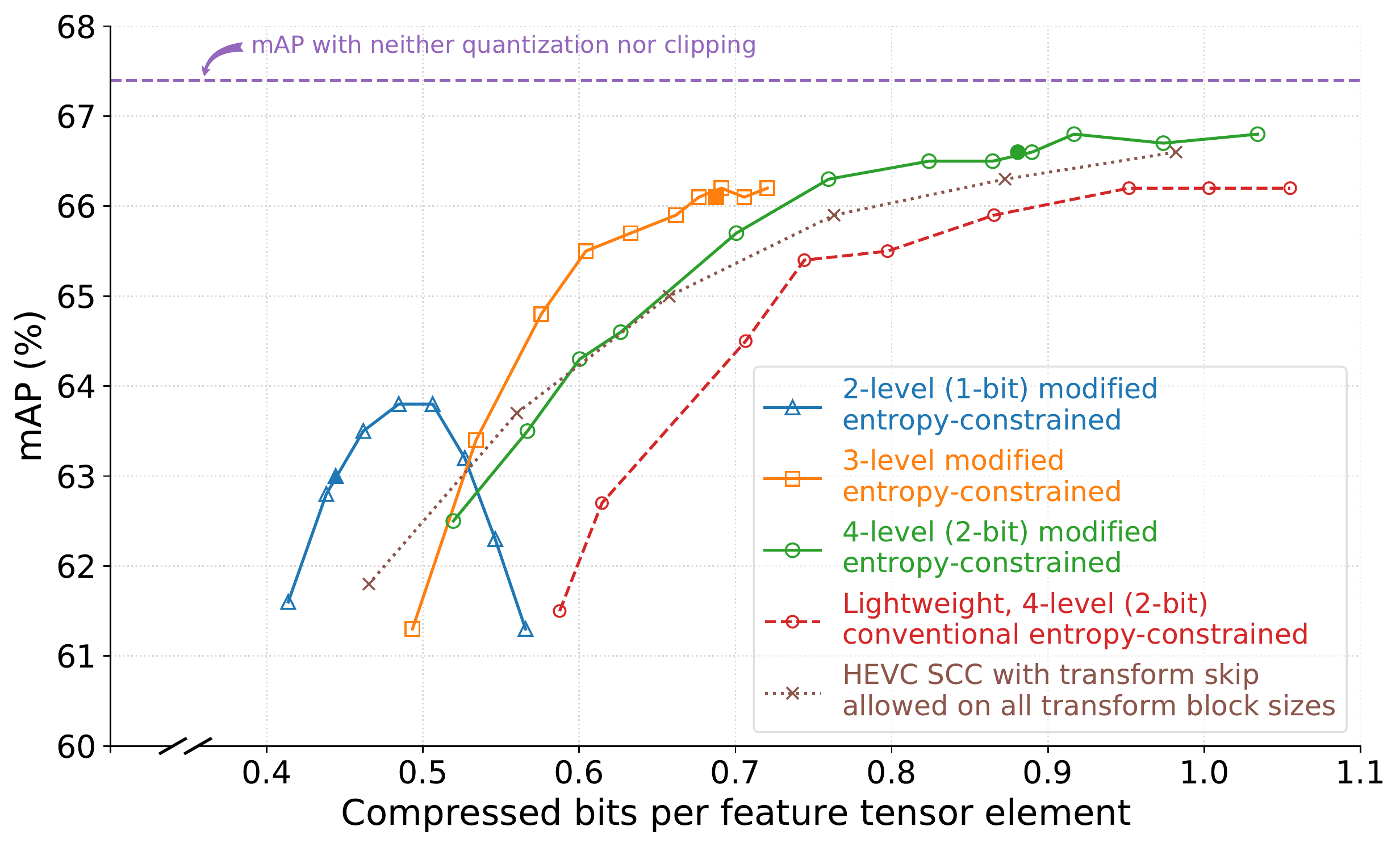}
    \caption{Object-detection mAP for YOLOv3 using lightweight compression at layer 12 with modified entropy-constrained quantization. Filled markers indicate corresponding performance using uniform quantization.}
    \label{fig:yolov3_plot}
\end{figure}

The performance of lightweight compression with modified entropy-constrained quantization is shown in Fig.~\ref{fig:resnet50_top1_plot} for \resnetfifty image classification, and in Fig.~\ref{fig:yolov3_plot} for YOLOv3 object detection. We show the performance here using extremely coarse quantization, namely 2--3 levels, which corresponds to a 1--2 bit quantized representation, followed by binarization and entropy coding to generate the final compressed bit-stream.
We also include for comparison the best-performing HEVC-SCC curves from Fig.~\ref{fig:resnet_yolo_RD}. We also show the performance when using an entropy-constrained quantizer designed using the conventional algorithm, which does not pin the outermost reconstruction levels. With 4-level quantization, using the modified quantizer design method improved the neural network performance by about 0.5--1.5\% as compared to when using the conventional algorithm. We can also see that the entropy-constrained quantizer's ability to cover a range of rates enabled us to improve the 2-level (1-bit) performance by about 1\% for YOLOv3. For \resnetfifty, the modified quantizer design algorithm also allowed us to obtain improved accuracies for 3- and 4-level (2-bit) quantizers. The range of achievable compressed bit-stream sizes for all these experiments with quantization to 4 and fewer levels was between about 0.3 to 1.0 bits per tensor element.


\section{Conclusions}
\label{sec:conclusions}
We presented an efficient and lightweight post-training compression method for coding the intermediate feature tensors of a split deep neural network. The codec only requires clipping, coarse quantization, binarization, and entropy coding to compress feature tensors, so is well over 90\% less complex than existing image or video codecs such as HEVC that are typically used for picture compression. We improved upon our earlier results by presenting an analytic model for obtaining optimal clipping ranges for feature tensors output by leaky ReLU activation functions. We used the new models with ReLU and leaky ReLU to estimate clipping and quantization error and showed that it produced a good match to empirically obtained results. We also presented an entropy-constrained quantizer design algorithm that pinned boundary reconstruction levels to quantize clipped activations, resulting in a 0.5--1.5\% improvement in performance as compared to using the conventional algorithm. With this lightweight lossy compression technique, we were able to quantize the 32-bit floating point activations output by a split DNN to fewer than 2 bits per element and then compress them further to 0.6 to 0.8 bits per element, while keeping the loss in output precision or accuracy to less than 1\%. We also showed that the lightweight codec yielded accuracies of up to 1.3\% higher than HEVC-SCC. The performance and simplicity of this lightweight compression technique makes it an attractive option for coding activations for edge/cloud DNN applications.


\bibliographystyle{IEEEbib-abbrev} 
\bibliography{IEEEabrv,strings,refs}

\begin{thebibliography}{10}

\bibitem{Chen2019}
J. {Chen} and X. {Ran},
\newblock ``Deep learning with edge computing: A review,''
\newblock {\em Proceedings of the IEEE}, vol. 107, no. 8, pp. 1655--1674, Aug.
  2019.

\bibitem{Lane2018}
N.~D. {Lane} and P. {Warden},
\newblock ``The deep (learning) transformation of mobile and embedded
  computing,''
\newblock {\em Computer}, vol. 51, no. 5, pp. 12--16, May 2018.

\bibitem{Tan_2019_CVPR}
M. Tan, B. Chen, R. Pang, V. Vasudevan, and Q.~V. Le,
\newblock ``{MnasNet}: Platform-aware neural architecture search for mobile,''
\newblock {\em 2019 IEEE/CVF Conf. on Computer Vision and Pattern Recognition
  (CVPR)}, pp. 2815--2823, June 2018.

\bibitem{Kang2017}
Y. Kang, J. Hauswald, C. Gao, A. Rovinski, T.~N. Mudge, J. Mars, and L. Tang,
\newblock ``Neurosurgeon: Collaborative intelligence between the cloud and
  mobile edge,''
\newblock in {\em ASPLOS '17}, Apr. 2017.

\bibitem{Bajic_etal_ICASSP21}
I.~V. Baji\'{c}, W. Lin, and Y. Tian,
\newblock ``Collaborative intelligence: Challenges and opportunities,''
\newblock in {\em Proc. IEEE ICASSP}, 2021,
\newblock to appear. Available: arXiv:2102.06841.

\bibitem{zhang2015efficient}
X. Zhang, J. Zou, X. Ming, K. He, and J. Sun,
\newblock ``Efficient and accurate approximations of nonlinear convolutional
  networks,''
\newblock in {\em 2015 IEEE/CVF Conf. Computer Vision and Pattern Recognition
  (CVPR)}, 2015, pp. 1984--1992.

\bibitem{zhuang2018discrimination}
Z. Zhuang, M. Tan, B. Zhuang, J. Liu, Y. Guo, Q. Wu, J. Huang, and J. Zhu,
\newblock ``Discrimination-aware channel pruning for deep neural networks,''
\newblock in {\em Deep Learning and Unsupervised Feature Learning Workshop,
  NIPS}, 2018, pp. 875--886.

\bibitem{Mishra2017_WRPN}
A. Mishra, E. Nurvitadhi, J.~J. Cook, and D. Marr,
\newblock ``{WRPN}: Wide reduced-precision networks,''
\newblock in {\em 6th Int. Conf. on Learning Representations (ICLR)}, May 2018.

\bibitem{jacob2018quantization}
B. Jacob, S. Kligys, B. Chen, M. Zhu, M. Tang, A. Howard, H. Adam, and D.
  Kalenichenko,
\newblock ``Quantization and training of neural networks for efficient
  integer-arithmetic-only inference,''
\newblock in {\em Proceedings of the IEEE Conf. Computer Vision and Pattern
  Recognition (CVPR)}, 2018, pp. 2704--2713.

\bibitem{Cohen_quantcode_ICME2020}
R.~A. Cohen, H. Choi, and I.~V. Baji\'{c},
\newblock ``Lightweight compression of neural network feature tensors for
  collaborative intelligence,''
\newblock {\em Proc. 21st IEEE Int. Conf. Multimedia and Expo (ICME)}, July
  2020.

\bibitem{Maas2013RectifierNI}
A.~L. Maas, A.~Y. Hannun, and A.~Y. Ng,
\newblock ``Rectifier nonlinearities improve neural network acoustic models,''
\newblock in {\em Int. Conf.Machine Learning (ICML)}, 2013.

\bibitem{He2015DeepRL}
K. He, X. Zhang, S. Ren, and J. Sun,
\newblock ``Deep residual learning for image recognition,''
\newblock {\em {IEEE} Conf. Computer Vision and Pattern Recognition ({CVPR})},
  pp. 770--778, June 2016.

\bibitem{Banner2018_8BitTraining}
R. Banner, I. Hubara, E. Hoffer, and D. Soudry,
\newblock ``Scalable methods for 8-bit training of neural networks,''
\newblock in {\em Proc. 32nd Int. Conf. Neural Information Processing Systems
  ({NeurIPS})}, Dec. 2018, pp. 5151--–5159.

\bibitem{Choi2019_2bit}
J. Choi, S. Venkataramani, V. Srinivasan, K. Gopalakrishnan, Z. Wang, and P.
  Chuang,
\newblock ``Accurate and efficient 2-bit quantized neural networks,''
\newblock in {\em Proc. 2\textsuperscript{nd} {SysML} Conf.}, Mar. 2019.

\bibitem{Hubara2016_BNN}
I. Hubara, M. Courbariaux, D. Soudry, R. El-Yaniv, and Y. Bengio,
\newblock ``Binarized neural networks,''
\newblock in {\em Proc. 30th Int. Conf. Neural Information Processing Systems
  ({NIPS})}, Dec. 2016, pp. 4114--–4122.

\bibitem{Rastegari2016_XNOR}
M. Rastegari, V. Ordonez, J. Redmon, and A. Farhadi,
\newblock ``{XNOR-Net}: {ImageNet} classification using binary convolutional
  neural networks,''
\newblock in {\em 14th European Conf. on Computer Vision ({ECCV})}, Oct. 2016,
  pp. 525--542.

\bibitem{Krishnamoorthi2018_QuantizingDC}
R. Krishnamoorthi,
\newblock ``Quantizing deep convolutional networks for efficient inference: A
  whitepaper,''
\newblock {\em arXiv abs/1806.08342}, June 2018.

\bibitem{Park2018a}
E. Park, D. Kim, and S. Yoo,
\newblock ``Energy-efficient neural network accelerator based on outlier-aware
  low-precision computation,''
\newblock in {\em 2018 ACM/IEEE 45th Annual Int. Symposium Computer
  Architecture (ISCA)}, 2018, pp. 688--698.

\bibitem{Park2018b}
E. Park, S. Yoo, and P. Vajda,
\newblock ``Value-aware quantization for training and inference of neural
  networks,''
\newblock in {\em 15th European Conf. on Computer Vision ({ECCV})}, Sep. 2018.

\bibitem{Zhao2019_Quantization}
R. Zhao, Y. Hu, J. Dotzel, C.~D. Sa, and Z. Zhang,
\newblock ``Improving neural network quantization without retraining using
  outlier channel splitting,''
\newblock in {\em Proc. 36th Int. Conf. on Machine Learning, {ICML} 2019}, June
  2019, pp. 7543--7552.

\bibitem{Nagel2019}
M. Nagel, M. van Baalen, T. Blankevoort, and M. Welling,
\newblock ``Data-free quantization through weight equalization and bias
  correction,''
\newblock in {\em 2019 IEEE Int. Conf. Computer Vision ({ICCV})}, 2019.

\bibitem{banner2018_ACIQ}
R. Banner, Y. Nahshan, E. Hoffer, and D. Soudry,
\newblock ``{ACIQ}: Analytical clipping for integer quantization of neural
  networks,'' [Online]: \url{https://openreview.net/forum?id=B1x33sC9KQ}, Sept.
  2018.

\bibitem{Banner2019_4bit}
R. Banner, Y. Nahshan, E. Hoffer, and D. Soudry,
\newblock ``Post-training 4-bit quantization of convolution networks for
  rapid-deployment,''
\newblock in {\em Proc. 33rd Int. Conf. Neural Information Processing Systems
  ({NeurIPS})}, May 2019, pp. 7950--7958.

\bibitem{Cai2019}
Y. Cai, Z. Yao, Z. Dong, A. Gholami, M.~W. Mahoney, and K. Keutzer,
\newblock ``{ZeroQ}: A novel zero shot quantization framework,''
\newblock in {\em {IEEE} Conf. Computer Vision and Pattern Recognition
  ({CVPR})}, June 2020.

\bibitem{dfc_for_collab_object_detection}
H. Choi and I.~V. Baji\'{c},
\newblock ``Deep feature compression for collaborative object detection,''
\newblock {\em Proc. 25th IEEE Int. Conf. Image Processing (ICIP)}, pp.
  3743--3747, Oct. 2018.

\bibitem{eshratifar2019jointdnn}
A.~E. Eshratifar, M.~S. Abrishami, and M. Pedram,
\newblock ``{JointDNN}: An efficient training and inference engine for
  intelligent mobile cloud computing services,''
\newblock {\em {IEEE} Trans. Mobile Comput.}, Oct. 2019.

\bibitem{Choi2018NearLosslessDF}
H. Choi and I.~V. Baji\'{c},
\newblock ``Near-lossless deep feature compression for collaborative
  intelligence,''
\newblock {\em IEEE 20th Int. Workshop on Multimedia Signal Processing (MMSP)},
  Aug. 2018.

\bibitem{eshratifar2019towards}
A.~E. Eshratifar, A. Esmaili, and M. Pedram,
\newblock ``Towards collaborative intelligence friendly architectures for deep
  learning,''
\newblock {\em 20th Int. Symposium Quality Electronic Design (ISQED)}, pp.
  14--19, Mar. 2019.

\bibitem{eshratifar2019bottlenet}
A.~E. {Eshratifar}, A. {Esmaili}, and M. {Pedram},
\newblock ``{BottleNet}: A deep learning architecture for intelligent mobile
  cloud computing services,''
\newblock in {\em Proc. IEEE/ACM Int. Symposium Low Power Electronics and
  Design (ISLPED)}, July 2019.

\bibitem{Choi_BaF_2020}
H. Choi, R.~A. Cohen, and I.~V. Baji\'{c},
\newblock ``Back-and-forth prediction for deep tensor compression,''
\newblock {\em Proc. 45th IEEE Int. Conf. Acoustics, Speech, and Signal
  Processing (ICASSP)}, May 2020,
\newblock in press.

\bibitem{Vanhoucke2011}
V. Vanhoucke, A. Senior, and M.~Z. Mao,
\newblock ``Improving the speed of neural networks on {CPUs},''
\newblock in {\em Deep Learning and Unsupervised Feature Learning Workshop,
  NIPS}, Dec. 2011.

\bibitem{imagenet2015}
O. Russakovsky et~al.,
\newblock ``{ImageNet Large Scale Visual Recognition Challenge},''
\newblock {\em Int. J. Comput. Vision}, vol. 115, no. 3, pp. 211--252, Dec.
  2015.

\bibitem{alexnet}
A. Krizhevsky, I. Sutskever, and G.~E. Hinton,
\newblock ``Imagenet classification with deep convolutional neural networks,''
\newblock in {\em Advances in Neural Information Processing Systems}. 2012,
  vol.~25, Curran Associates, Inc.

\bibitem{Redmon2018_yolov3}
J. Redmon and A. Farhadi,
\newblock ``{YOLOv3}: An incremental improvement,''
\newblock {\em arXiv preprint arXiv:1804.02767}, Apr. 2018.

\bibitem{COCO}
T.-Y. Lin, M. Maire, S. Belongie, J. Hays, P. Perona, D. Ramanan, P.
  Doll{\'a}r, and C.~L. Zitnick,
\newblock ``Microsoft {COCO}: Common objects in context,''
\newblock in {\em European Conf. on Computer Vision (ECCV)}, Sept. 2014.

\bibitem{Kozubowski2000}
T. Kozubowski and K. Podgorski,
\newblock ``A multivariate and asymmetric generalization of {L}aplace
  distribution,''
\newblock {\em Computational Statistics}, vol. 15, pp. 531--540, Dec. 2000.

\bibitem{Chou1989}
P.~A. {Chou}, T. {Lookabaugh}, and R.~M. {Gray},
\newblock ``Entropy-constrained vector quantization,''
\newblock {\em {IEEE} Trans. Acoust., Speech, Signal Process.}, vol. 37, no. 1,
  pp. 31--42, Jan. 1989.

\bibitem{Girod_ecquant}
B. Girod,
\newblock ``Quantization,'' {EE398A} Image and Video Compression, [Online]:
  \url{https://web.stanford.edu/class/ee398a/handouts/lectures/05-Quantization.pdf},
\newblock Accessed: 2020-04-02.

\bibitem{Marpe2003_CABAC}
D. Marpe, H. Schwarz, and T. Wiegand,
\newblock ``Context-based adaptive binary arithmetic coding in the {H.264/AVC}
  video compression standard,''
\newblock {\em {IEEE} Trans. Circuits Syst. Video Technol.}, vol. 13, no. 7,
  pp. 620--636, July 2003.

\bibitem{Bossen1012_hevc_complexity}
F. Bossen, B. Bross, K. Suhring, and D. Flynn,
\newblock ``{HEVC} complexity and implementation analysis,''
\newblock {\em {IEEE} Trans. Circuits Syst. Video Technol.}, vol. 22, no. 12,
  pp. 1685--1696, Dec. 2012.

\bibitem{darknet_weights}
J. Redmon,
\newblock ``Darknet: {O}pen source neural networks in {C},'' [Online]:
  \url{https://pjreddie.com/darknet},
\newblock Accessed: 2021-02-25.

\bibitem{AlexeyAB_darknet}
A. Bochkovskiy,
\newblock ``{darknet},'' [Online]:
  \url{https://github.com/AlexeyAB/darknet/tree/8c80ba6},
\newblock Accessed: 2020-11-22.

\bibitem{COCO_API}
``{COCO API},'' [Online]: \url{https://github.com/cocodataset/cocoapi},
\newblock Accessed: 2019-03-19.

\bibitem{HM16.20}
``{HEVC} reference software ({HM} 16.20),'' [Online]:
  \url{http://hevc.hhi.fraunhofer.de/svn/svn_HEVCSoftware/tags/HM-16.20+SCM-8.8},
\newblock Accessed: 2019-12-12.

\bibitem{hevc_std_2017}
``High efficiency video coding,'' {ITU-T and ISO/IEC},
\newblock Rec. ITU-T H.265 \textbar{} ISO/IEC 23008-2:2017, 2017.

\end{thebibliography}



\begin{IEEEbiography}[{\includegraphics[width=1in,height
=1.25in,clip,keepaspectratio]{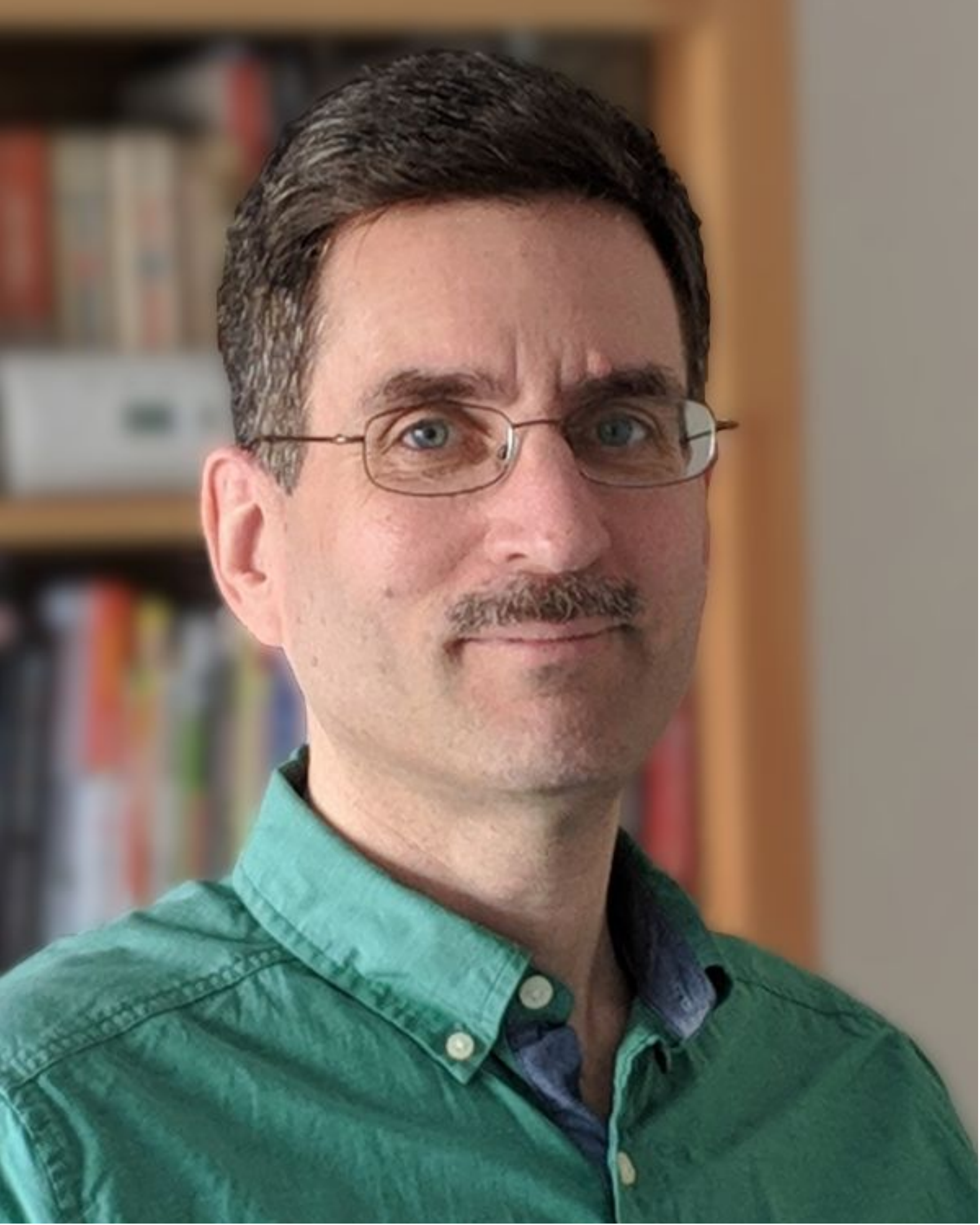}}]{Robert A Cohen} (S’85--M'90--SM’12) received the B.S. and M.Eng. degrees in computer and systems engineering and the Ph.D. degree in electrical engineering from Rensselaer Polytechnic Institute, Troy, USA. He held positions with IBM, San Jose, CA, Harris RF Communications, Rochester, NY, and with Philips Research, Briarcliff Manor, NY, USA, where he performed research in areas related to the Grand Alliance HDTV decoder, rapid prototyping for very large-scale integration video systems, statistical multiplexing for digital video encoders, scalable MPEG-4 video streaming, and next-generation video surveillance systems. He was a Principal Member of the Research Staff with Mitsubishi Electric Research Laboratories, Cambridge, MA, USA until 2018, where he researched next-generation video coding, HEVC screen-content coding, perceptual image/video coding, and 3D point cloud compression. He also actively participated in ISO/MPEG and ITU standardization activities, including chairing several ad hoc groups and core experiments, contributing to High Efficiency Video Coding-related call for proposals, drafting the Joint Call for Proposals for coding of screen content in JCT-VC, and editing the ISO/IEC working draft for geometry-based point cloud compression. He is currently a Senior Research Scientist at Simon Fraser University, Burnaby, BC, Canada, where he performs research on deep learning and compression. His interests include deep learning, video coding and communications, video, image, and signal processing, digital systems, and point cloud compression. Dr. Cohen organized the Special Session on Screen Content Coding in PCS 2013, and was a member of the IEEE Signal Processing Society Industrial Relations Committee. He was also a Guest Editor of Signal Processing: Image Communication of the Special Issue on Advances in High Dynamic Range Video Research.
\end{IEEEbiography}


\begin{IEEEbiography}[{\includegraphics[width=1in,height
=1.25in,clip,keepaspectratio]{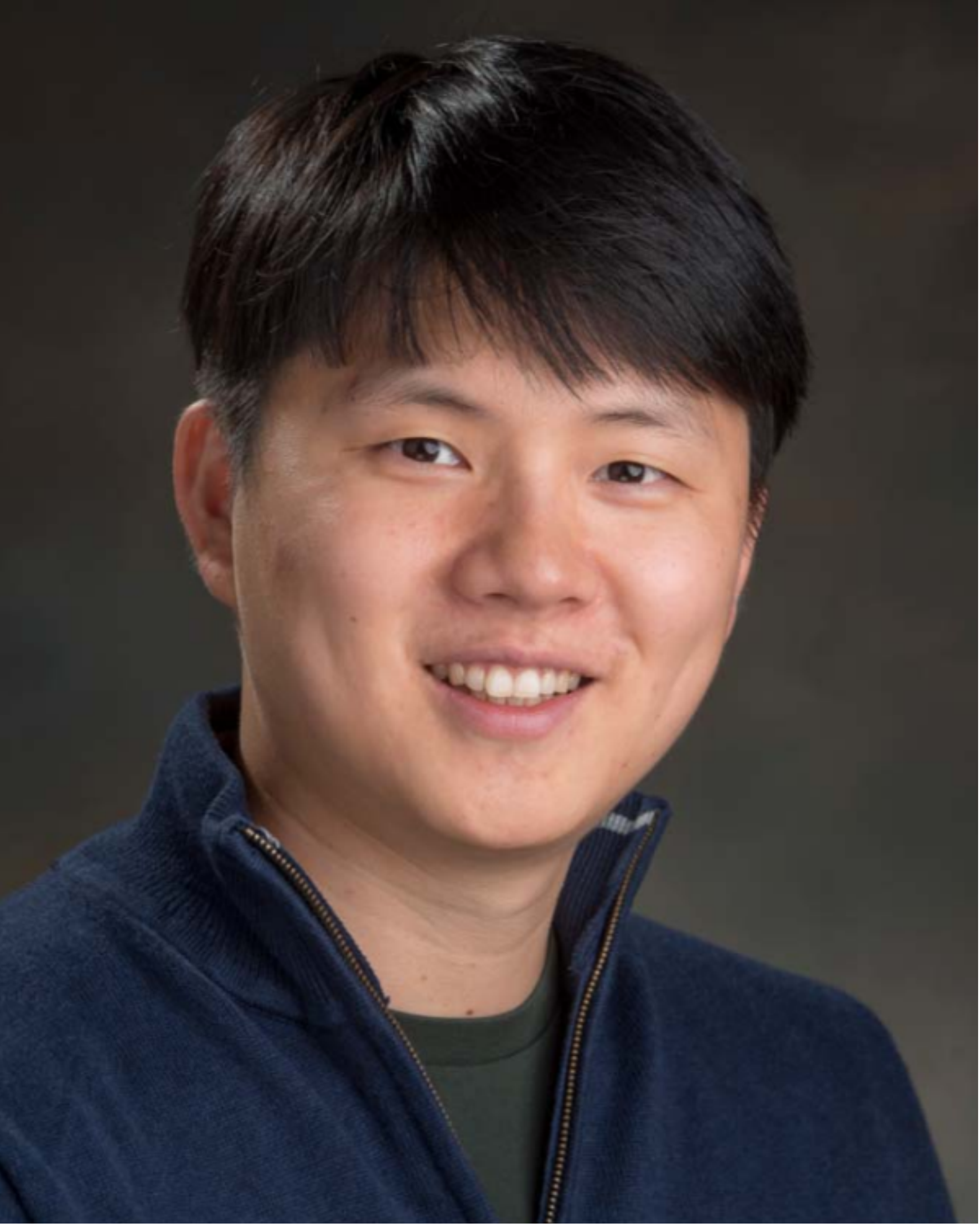}}]{Hyomin Choi}(S’17) is a Ph.D. student at Simon Fraser University, Burnaby, BC, Canada. He received B.S. and M.S. degrees in computer engineering from Kwangwoon University, Seoul, Korea, in 2010 and 2012, respectively. He was a research engineer at System IC Research Center, LG Electronics, from 2012 to 2016. His research interests include image/video coding and deep learning. He was the winner of the Vanier Canada Graduate Scholarships (Vanier CGS) in 2017. He was the recipient of the IEEE SPS Student Travel Grant for ICIP 2018 and a finalist in the Asilomar 2020 Student Paper Competition.
\end{IEEEbiography}

\vskip -12pt plus -1fil

\begin{IEEEbiography}[{\includegraphics[width=1in,height
=1.25in,clip,keepaspectratio]{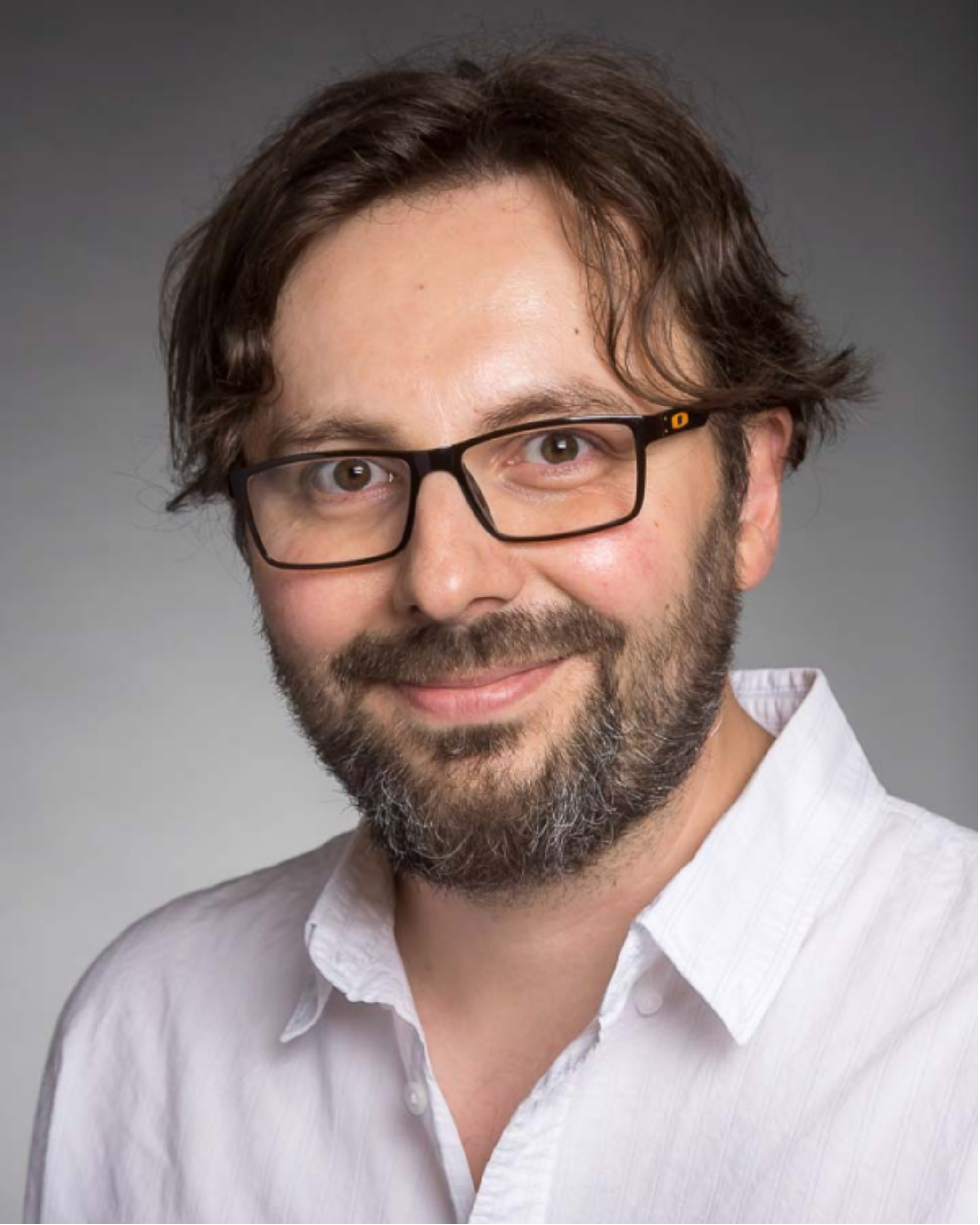}}]{Ivan V. Baji\'{c}}(S'99--M’04--SM’11) received the Ph.D. degree in electrical engineering from Rensselaer Polytechnic Institute, Troy, NY, USA, in 2003. He is a Professor of Engineering Science and co-director of the Multimedia Lab at Simon Fraser University, Burnaby, BC, Canada. His research interests include signal processing and machine learning with applications to multimedia signal processing, compression, communications, and collaborative intelligence. He has authored about a dozen and co-authored another ten dozen publications in these fields. He is currently the Vice Chair of the IEEE Multimedia Signal Processing Technical Committee and an elected member of the IEEE Multimedia Systems and Applications Technical Committee. He has served on the organizing and/or program committees of the main conferences in the field. He was an Associate Editor of IEEE Transactions on Multimedia and IEEE Signal Processing Magazine, and is currently a Senior Area Editor of IEEE Signal Processing Letters.
\end{IEEEbiography}

\end{document}